\documentclass[conference]{IEEEtran}
\pdfoutput=1
\usepackage{tikz}
\usetikzlibrary{shapes,snakes}
\usepackage{amsmath}
\usepackage{amssymb}
\usepackage{flushend}
\usepackage{multirow}
\usepackage{bclogo}
\usepackage{hyperref}
\usepackage{graphicx}
\usepackage{subfiles}
\usetikzlibrary{patterns, matrix}

\ifCLASSINFOpdf
\else
\fi
\begin{document}
%
\title{Spiking Approximations of the MaxPooling Operation in Deep SNNs}

\author{\IEEEauthorblockN{Ramashish Gaurav}
\IEEEauthorblockA{Systems Design Engineering\\
University of Waterloo, Canada\\
Email: rgaurav@uwaterloo.ca}
\and
\IEEEauthorblockN{Bryan Tripp}
\IEEEauthorblockA{Systems Design Engineering\\
University of Waterloo, Canada\\
Email: bptripp@uwaterloo.ca}
\and
\IEEEauthorblockN{Apurva Narayan}
\IEEEauthorblockA{Department of Computer Science\\
University of British Columbia, Canada\\
Email: apurva.narayan@ubc.ca}}


%


\maketitle

\begin{abstract}
Spiking  Neural  Networks  (SNNs)  are  an  emerging  domain of  biologically  inspired  neural  networks  that  have  shown promise for low-power AI. A number of methods exist for building deep SNNs, with Artificial Neural Network (ANN)-to-SNN conversion being highly successful. MaxPooling layers in Convolutional Neural Networks (CNNs) are an integral component to downsample the intermediate feature maps and introduce translational invariance, but the absence of their hardware-friendly spiking  equivalents limits such CNNs’ conversion to deep SNNs. In this paper, we present two hardware-friendly methods to implement Max-Pooling  in  deep  SNNs,  thus  facilitating  easy  conversion  of CNNs with MaxPooling layers to SNNs. In a first, we also execute SNNs with spiking-MaxPooling layers on Intel’s Loihi neuromorphic hardware (with MNIST, FMNIST, \& CIFAR10 dataset); thus, showing the feasibility of our approach.
\end{abstract}


%
\IEEEpeerreviewmaketitle

\section{Introduction}
Artificial Neural Networks (ANNs) have established themselves as the de-facto tool for a variety of Artificial Intelligence (AI) tasks. And the flagship performance of CNNs for image recognition/classification has remained unparalleled so far. However, their limitations in the aspects of energy consumption, robustness against noisy inputs, etc. has attracted interest in developing their spiking counterparts. Spiking Neural Networks (SNNs) offer a promise of low power AI and have shown to be more robust against noisy inputs \cite{stromatias2015robustness}, perturbations to the weights \cite{9207019}, and adversarial attacks \cite{8851732, sharmin2020inherent}. Out of a number of ways to build SNNs \cite{deep_snns}, the ANN-to-SNN conversion method has been highly effective for building deep SNNs. In this method, one first trains an ANN with traditional rate neurons (e.g. \texttt{ReLU}) and then replaces those rate neurons with spiking neurons, along with the other required modifications of weights \cite{sengupta2019going}, etc. For our work, we consider this ANN-to-SNN conversion paradigm to build deep SNNs.

MaxPooling in CNNs is a common method to downsample the intermediate feature maps obtained from the Convolutional layers. One can also use AveragePooling or Strided Convolution to downsample the feature maps; however, the choice of the pooling method is contextual \cite{boureau2010theoretical}, and MaxPooling is generally found to give better performance. That said, \cite{boureau2010theoretical} also found that ``depending on the data and features, either max or average pooling may perform best''. Several architectures e.g. ResNet-$50$ (-152) \cite{he2016deep}, Xception \cite{chollet2017xception}, EfficientNet \cite{tan2019efficientnet} which form the backbone of different methods to achieve SoTA results on the ImageNet, use a mix of Max and AveragePooling layers (or GlobalMax/Average Pooling layers).

However, the conversion of CNNs with MaxPooling layers to SNNs is a convoluted process. MaxPooling in SNNs has been a long-standing problem; only a handful of approaches exist for the same. \cite{hu2016max} present three approaches for MaxPooling in SNNs; where they design a pooling gate to monitor the spiking neurons' activities (in a pooling window) and dynamically connect one of the neurons to the output neuron based on their criteria for the maximally firing neuron. Such a gating mechanism is leveraged by \cite{rueckauer2017conversion} too, where they employ a finite impulse response filter to control the gating function and do MaxPooling in SNNs; \cite{kim2020spiking} too use the same. Few other works \cite{masquelier2007unsupervised, zhao_wta, orchard2015hfirst, li2017bio, mozafari2019spyketorch} leverage a Time-To-First-Spike based temporal Winner Take All (WTA) mechanism (or its variants) to do MaxPooling in SNNs; where the earliest occurring spike in a pooling area is sent to the next layer, with rest of the neurons (in the pooling area) reset to blocked. Similarly, \cite{lin2017quantisation} employ a lateral inhibition based method to do MaxPooling in SNNs. In another novel approach, \cite{nguyen2020lightweight} select a neuron (in the pooling region) with the highest membrane potential to output the spikes; they do a soft reset (i.e. reset by subtraction) of the firing neuron's membrane potential. They also propose a hardware architecture for their method of MaxPooling. However, \textit{none of the above methods have been evaluated on a specialised neuromorphic hardware} (e.g. Loihi \cite{davies2018loihi}), with the exception of WTA (but on FPGA \cite{orchard2015hfirst}).

The conversion of CNNs with AveragePooling layers to SNNs is trivial, as the AveragePooling layers can be modeled as convolution operation in the SNNs. Thus, a number of works replace the MaxPooling layers with AveragePooling layers \cite{wu2018spatio, sengupta2019going, cheng2020lisnn, garg2020dct, kucik2021investigating, kundu2021spike, yan2021near, huang2021fpga} or with Strided Convolutional layers \cite{esser2016convolutional, patel2021spiking} in their network; often leading to weaker ANNs \cite{rueckauer2017conversion}. Overall, this dearth of spiking-MaxPooling approaches in SNNs and \textit{the neuromorphic hardware-unfriendliness of the existing ones} motivated us to build spiking equivalents of the MaxPooling operation \textit{which can be entirely deployed on a neuromorphic hardware} (e.g. Loihi). Our contributions are outlined below:

\begin{itemize}
    \item We propose two methods to do MaxPooling in SNNs, and evaluate them on the Loihi neuromorphic chip
    \item In a first, we also deploy deep SNNs with MaxPooling layers on the Loihi boards with one of our methods
\end{itemize}

We next present our \textbf{Methods} of spiking-MaxPooling, followed by the \textbf{Experiments \& Results} section, and a \textbf{Discussion} on the result analysis and adaptability of our methods.

\section{Methods}
In this section we present our proposed methods of spiking-MaxPooling. We start with each method's elemental details, followed by their Proof-of-Concept Demonstration on the Loihi chip. Henceforth, otherwise stated, all the instances of ``neurons" are Integrate \& Fire (IF) spiking neurons.

\subsection{Method 1: MAX join-Op}
MAX join-Op (MJOP) based spiking-MaxPooling leverages the low level NxCore APIs made available through the NxSDK tool (to program the Loihi chips) by Intel; thus, this method is Loihi hardware dependent. A single Loihi chip consists of $128$ Neuro-Cores, where each Neuro-Core implements $1024$ Single Compartment (SC) spiking units. Each compartment can be simulated as an individual neuron or can become part of a Multi-Compartment (MC) neuron. Each MC neuron is a binary tree of single compartments, where each node (a compartment) can have at most two child nodes/dendrites (also compartments). Note that a MC neuron cannot span across two or more Neuro-Cores; rather is limited to just one Neuro-Core. Thus, a MC neuron can have a maximum of $1024$ SC. Through each connection between the compartments (in a MC neuron), the two state variables: current ($U$) and voltage ($V$) can flow. Thus, a compartment can communicate either its $U$ or $V$ to its parent compartment. The parent compartment incorporates the state variables from its dendrites with its own $U$ by following a \textit{join} operation defined for its dendritic connections. Note that, an external $U$ can be injected to each compartment (including the root compartment) in a MC neuron, thereby enabling them to spike (provided their $V$ reaches the threshold). Also, note that the neurons (MC or SC) in a Loihi chip communicate via spikes only; they cannot exchange $U$ or $V$ directly, which falls in line with the biological neurons.

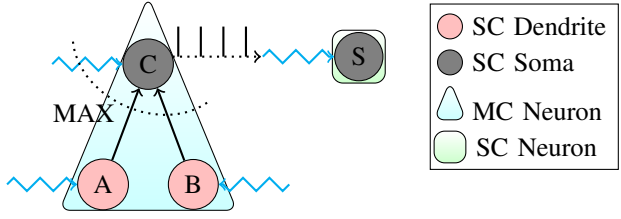
\begin{figure}[]
\begin{tikzpicture}[
scale=2, 
d_neuron/.style={draw, circle, fill=pink},
s_neuron/.style={draw, circle, fill=gray},
multi_compartment/.style={draw, isosceles triangle, rounded corners, top color = white, bottom color = cyan!20, rotate=90},
single_compartment/.style={draw, rectangle, rounded corners, top color = white, bottom color = green!20},
]

\node[multi_compartment, scale=5.5] at (0.4, 0) (A){};
\node[s_neuron] at (0.4, 0.65) (B){C};
\node[d_neuron] at (0.1, -0.15) (C){A};
\node[d_neuron] at (0.7, -0.15) (D){B};
\node[single_compartment, scale=3] at (1.8, 0.7) (J){};
\node[s_neuron] at (1.8, 0.7) (K){S};

\node[] at (-0.6, -0.15) (E){};
\node[] at (-0.3, 0.65) (F){};
\node[] at (1.4, -0.15) (G){};
\node[] at (0.5, 0.7) (H){};
\node[] at (1.1, 0.7) (I){};

\path[->] (C) edge [thick] (B);
\path[->] (D) edge [thick] (B);

\path[->, draw=cyan, snake, thick] (E) -> (C);
\path[->, draw=cyan, snake, thick] (F) -> (B);
\path[->, draw=cyan, snake, thick] (G) -> (D);
\path[->, draw=cyan, snake, thick] (I) -> (K);
\path[->, draw, dotted, thick] (H) -> (1.15, 0.7);

\path[-] (-0.1, 0.8) edge [dotted, thick, bend left = -60] node[below=0.1cm, left=0.01cm] {MAX} (0.8, 0.4);

\draw[thick] (0.6, 0.7) -> (0.6, 0.9);
\draw[thick] (0.75, 0.7) -> (0.75, 0.9);
\draw[thick] (0.9, 0.7) -> (0.9, 0.9);
\draw[thick] (1.05, 0.7) -> (1.05, 0.9);

\node[draw, rectangle, minimum width=2.5cm, minimum height = 2.2cm] at (2.9, 0.5) (LEGEND) {};
\matrix [] at (2.9, 0.5) {
  \node [d_neuron,label=right:SC Dendrite]{}; \\
  \node [s_neuron,label=right:SC Soma]{}; \\
  \node [multi_compartment, label=below:MC Neuron]{}; \\
  \node [single_compartment, scale=1.8, label=right:SC Neuron]{};\\
};

\end{tikzpicture}
\caption{SC: Single Compartment, MC: Multi-Compartment. Wiggly arrows denote post-synaptic current due to the incoming spikes; Dotted straight arrow shows the resulting spike-train due to \texttt{C}'s spiking activity; Dotted curved line over $3$ arrows shows the $3$ join-Op arguments.}
\label{fig:mc_neuron}
\end{figure}

Fig. \ref{fig:mc_neuron} shows an example of a MC neuron communicating its spikes to a SC neuron. The MC neuron has a binary tree structure, with the root node (i.e. soma compartment \texttt{C}) having two child leaf nodes (dendrite compartments \texttt{A} \& \texttt{B}). In Loihi, each compartment's voltage at time-step $t$ i.e. $X.V[t]$ ($X$ can be a soma or its dendrites) is updated by the following rule:
\begin{equation}
    X.V[t] = X.V[t-1] \times (1-decay) + X.dV[t]
\end{equation}
where the value of $X.dV[t]$ is found by applying the specific \textit{join} operation (join-Op) e.g. ADD, MAX, MIN, etc. over its current $X.U[t]$ and the input state variables from its dendrites.  We leverage this MC neuron creation functionality and the MAX join-Op for realizing spiking-MaxPooling (on Loihi boards) in deep SNNs. We next explain the MAX join-Op in the context of the MC neuron in Fig. \ref{fig:mc_neuron}. For the same, we first define few terms: $X.U[t]$ and $X.bias$ denote the input current and bias current (respectively) of a compartment $X$. $X.U'[t]$ is the sum of $X.U[t]$ and $X.bias$. $A[t]$ (and $B[t]$) is the output of the child dendrite \texttt{A} (and \texttt{B}), which can either be $A.V[t]$ or $A.U'[t]$ (and $B.V[t]$ or $B.U'[t]$) depending upon which state variable we want to work with; we choose $U$. NxSDK defines the MAX join-Op as ($X$ being \texttt{C} here):
\begin{equation}
    C.dV[t] = max(C.U'[t], A[t], B[t])
    \label{eq:max_jop}
\end{equation}

We expand and simplify the Eq. \ref{eq:max_jop} by assuming $X.bias=0$ for $X \in \{\texttt{C}, \texttt{A}, \texttt{B}\}$ and setting the child compartments \texttt{A} and \texttt{B} to output current instead of $V$ (to its parent \texttt{C}). Thus, $A[t]=A.U'[t]$ and $B[t]=B.U'[t]$. Eq. \ref{eq:max_jop} simplifies as:
\begin{align}
    C.dV[t] &= max(C.U'[t], A[t], B[t])\\
            &= max(C.U[t] + C.bias, A[t], B[t])\\
            &= max(C.U[t], A.U'[t], B.U'[t])\\
            &= max(C.U[t], A.U[t], B.U[t])
\end{align}
Thus, in MJOP case, the root compartment (i.e. soma) \texttt{C}'s voltage dynamics is governed by the maximum of the input currents to it (from external source and its dendrites \texttt{A} \& \texttt{B}).

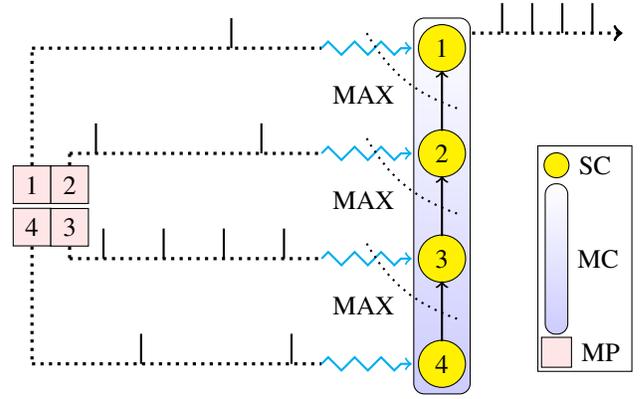
\begin{figure}
\begin{tikzpicture}[
scale=2, 
neuron/.style={draw, circle, fill=yellow},
multi_compartment/.style={draw, rectangle, minimum width=2cm, minimum height = 0.3cm, rounded corners, top color = white, bottom color = blue!20},
array/.style={matrix of nodes, nodes={draw, minimum size=5mm, fill=red!10},column sep=-\pgflinewidth, row sep=0.5mm, nodes in empty cells},
]

\matrix[array] at (0, -1.15) (pw) {1&2\\
                                4&3\\};

\node[multi_compartment, scale=2.5, rotate=90] at (2.6, -1.15) (A){};
\node[neuron] at (2.6, -0.1) (B){1};
\node[neuron] at (2.6, -0.8) (C){2};
\node[neuron] at (2.6, -1.5) (D){3};
\node[neuron] at (2.6, -2.2) (E){4};

\draw[dotted, very thick] (pw-1-1.north) to (-0.125, -0.1);
\draw[dotted, very thick] (-0.125, -0.1) -- (1.8, -0.1);
\path[->, draw=cyan, snake, thick] (1.8, -0.1) -> (2.4, -0.1);
\draw[thick] (1.2, -0.1) -> (1.2, 0.1);

\draw[dotted, very thick] (pw-1-2.north) to (0.125, -0.8);
\draw[dotted, very thick] (0.125, -0.8) -- (1.8, -0.8);
\path[->, draw=cyan, snake, thick] (1.8, -0.8) -> (2.4, -0.8);
\draw[thick] (0.3, -0.8) -> (0.3, -0.6);
\draw[thick] (1.4, -0.8) -> (1.4, -0.6);

\draw[dotted, very thick] (pw-2-2.south) to (0.125, -1.5);
\draw[dotted, very thick] (0.125, -1.5) -- (1.8, -1.5);
\path[->, draw=cyan, snake, thick] (1.8, -1.5) -> (2.4, -1.5);
\draw[thick] (0.35, -1.5) -> (0.35, -1.3);
\draw[thick] (0.75, -1.5) -> (0.75, -1.3);
\draw[thick] (1.15, -1.5) -> (1.15, -1.3);
\draw[thick] (1.55, -1.5) -> (1.55, -1.3);
\draw[dotted, very thick] (pw-2-1.south) to (-0.125, -2.2);
\draw[dotted, very thick] (-0.125, -2.2) -- (1.8, -2.2);
\path[->, draw=cyan, snake, thick] (1.8, -2.2) -> (2.4, -2.2);
\draw[thick] (0.6, -2.2) -> (0.6, -2.0);
\draw[thick] (1.6, -2.2) -> (1.6, -2.0);

\draw[thick] (3, 0) -> (3, 0.2);
\draw[thick] (3.2, 0) -> (3.2, 0.2);
\draw[thick] (3.4, 0) -> (3.4, 0.2);
\draw[thick] (3.6, 0) -> (3.6, 0.2);

\path[->, draw, thick] (2.6, -2.05) -- (2.6, -1.65);
\path[->, draw, thick] (2.6, -1.35) -- (2.6, -0.95);
\path[->, draw, thick] (2.6, -0.65) -- (2.6, -0.25);
\path[->, draw, dotted, very thick] (2.8, 0) -> (3.8, 0);

\path[-] (2.1, 0) edge [dotted, thick, bend left = -20] node[below=0.2cm, left=0.0cm] {MAX} (2.7, -0.5);

\path[-] (2.1, -0.7) edge [dotted, thick, bend left = -20] node[below=0.2cm, left=0.0cm] {MAX} (2.7, -1.2);

\path[-] (2.1, -1.4) edge [dotted, thick, bend left = -20] node[below=0.2cm, left=0.0cm] {MAX} (2.7, -1.9);


\node[draw, rectangle, minimum width=1.25cm, minimum height = 3cm] at (3.55, -1.5) (LEGEND) {};
\matrix [] at (3.55, -1.5) {
  \node [neuron,label=right:SC]{}; \\
  \node [multi_compartment, rotate=90,
  label=below:MC]{}; \\
  \node [draw, scale=1.7, fill=red!10,label=right:MP]{};\\
};
\end{tikzpicture}
\caption{\textit{MJOP Net}. SC: Single Compartment, MC: Multi-Compartment Neuron, MP: $2\times2$ MaxPooling Window. Integers $(1, 2, 3, 4)$ simply show a one-to-one correspondence. The incoming spikes (to the MC neuron) get synapsed/filtered to be fed as current $U$ (wiggly arrows) to each compartment.}
\label{fig:mc_neuron1}
\end{figure}

\subsubsection{MJOP Net for MaxPooling: }
In a conventional CNN architecture, MaxPooling is done over the activations of the preceding Convolutional layer rate-neurons. In an SNN, we can represent those real-valued activations (of rate-neurons) by passing the corresponding spiking-neurons' spike-trains through a low-pass filter (also known as filtering/synapsing). In other words, the synapsed spikes i.e. the post-synaptic current $U$ represents the activation. We leverage this characteristic of the SNNs and feed the individual currents $U_i$ (in a pooling window) to a MAX join-Op configured MC neuron. Note that the number of compartments in the MC neuron should be equal to the size of the pooling window.

For a $2\times2$ MaxPooling window, we construct a MC neuron with $4$ compartments as shown in the Fig. \ref{fig:mc_neuron1}. The outgoing spikes from each neuron in a pooling window induce a post-synaptic current $U$ in the respective individual compartments ($bias$ current of each compartment is set to 0). Each of the compartments (except the root/soma) is set to communicate its $U$ to its parent. The leaf node/compartment $4$ upon receiving the post-synaptic current updates its $V$ and communicates the received $U$ to its parent $3$. Compartment $3$ then computes the MAX of the current from its child $4$ and the incoming post-synaptic current, updates its $V$, and communicates the resulting maximum $U$ to its parent $2$. Compartments $2$ and $1$ repeat this same process; except that the compartment $1$ has no parent to communicate the MAX $U$ so far. The soma compartment $1$ then spikes at a rate \textit{corresponding} to the maximum input post-synaptic $U$ depending on its configuration, instead of outputting the max $U$. Note that in this network, MAX join-Op is executed over two arguments instead of three.
\subsubsection{Proof-of-Concept Demonstration: }
In MJOP Net, a running MAX of input currents $U_i$ is maintained, which is finally fed to the root/soma compartment. Mathematically:
\begin{equation}
    U_{out} = F(G(max(U_1, max(U_2, max(U_3, U_4)))))
    \label{eq:max_jop_out}
\end{equation}
where $U_i$ is the input current to compartment $i$, $G$ is the non-linear dynamical function of $U$ governing the voltage dynamics (thus, the spiking output) of soma, $F$ is the synaptic filter applied on the spike outputs of soma. Since the soma does not communicate the computed max current, rather its spikes to the next neuron (if connected), there arises a need to properly $scale$ the synapsed spikes i.e. $U_{out}$ to match ``True Max $U$'' ($=max(U_1, U_2, U_3, U_4)$ computed without spiking neurons on a non-neuromorphic hardware). We next execute the MJOP Net (in Fig. \ref{fig:mc_neuron1}) on the Loihi chip. The example MJOP net consists of a MAX join-Op configured MC neuron of $4$ compartments receiving periodic spiking inputs (spike amplitude $1$, with time-periods of $10$, $8$, $4$, and $6$ - a possible case when receiving the spike outputs of pooled neurons in a preceding Convolutional layer in an SNN). We set a probe on soma (compartment $1$) and filter its output spikes. In Fig. \ref{fig:max_jop_poc}a we see that the ``Scaled $U_{out}$'' matches the ``True Max $U$'' closely; the ``Average $U$'' is lower than the estimated max.

\subsection{Method 2: Absolute Value based Associative Max}
The representation of the real-valued activations of the rate-neurons as currents in SNNs inspires our second method as well. The Absolute Value based Associative Max (AVAM) method of spiking-MaxPooling is hardware independent and leverages the following two properties of the $max$ function.
\begin{equation}
    max(a, b) = \frac{a+b}{2} + \frac{|a-b|}{2}
    \label{eq:avam_1}
\end{equation}
\begin{equation}
\begin{split}
  max(a_1, a_2, a_3, \cdots, a_n) &= max(max(a_1, a_2), \\ & \cdots, max(a_{n-1}, a_n))
\label{eq:avam_2}
\end{split}
\end{equation}
where $a_i, a, b \in \mathbb{R}$, $n \in \mathbb{N}$; Eq. \ref{eq:avam_2} holds due to the Associative Property of $max()$. In Eq. \ref{eq:avam_1}, the average term $\frac{a+b}{2}$ (a linear operation) can be easily calculated via the connection weights from the neurons representing those values (i.e. $a$ and $b$); the challenge is to calculate the non-linear absolute value  function i.e. $|.|$. One can use the Neural Engineering Framework (NEF) principles \cite{eliasmith2003neural} to estimate the $|.|$ function by employing a network of \textit{Ensemble}s of neurons. However, the number of neurons in each \textit{Ensemble} can be large ($100s$ or more); thus, this method is not scalable. For the same reasons, a direct calculation of the $max()$ using NEF principles is not desirable.
\subsubsection{Estimation of $|.|$ function: } 
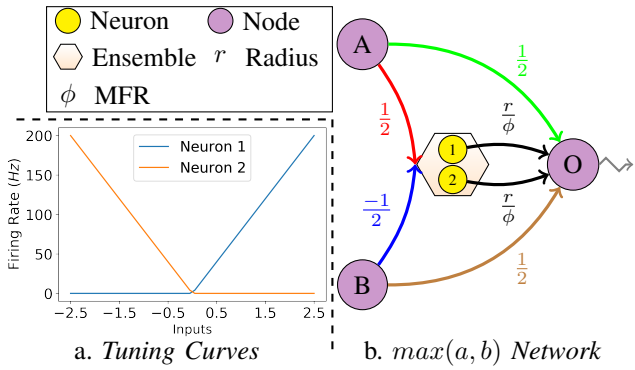
\begin{figure}[]
\begin{tikzpicture}[
scale=2, 
neuron/.style={draw, circle, fill=yellow},
ens_neuron/.style={draw, circle, fill=yellow, scale=0.6},
ens/.style={draw, regular polygon, regular polygon sides=6, scale=1, top color =white , bottom color = orange!20},
node/.style={draw, circle, fill=violet!40},
]
\node[] at (-0.5, -0.75) (tc_pic) {\includegraphics[scale=0.3]{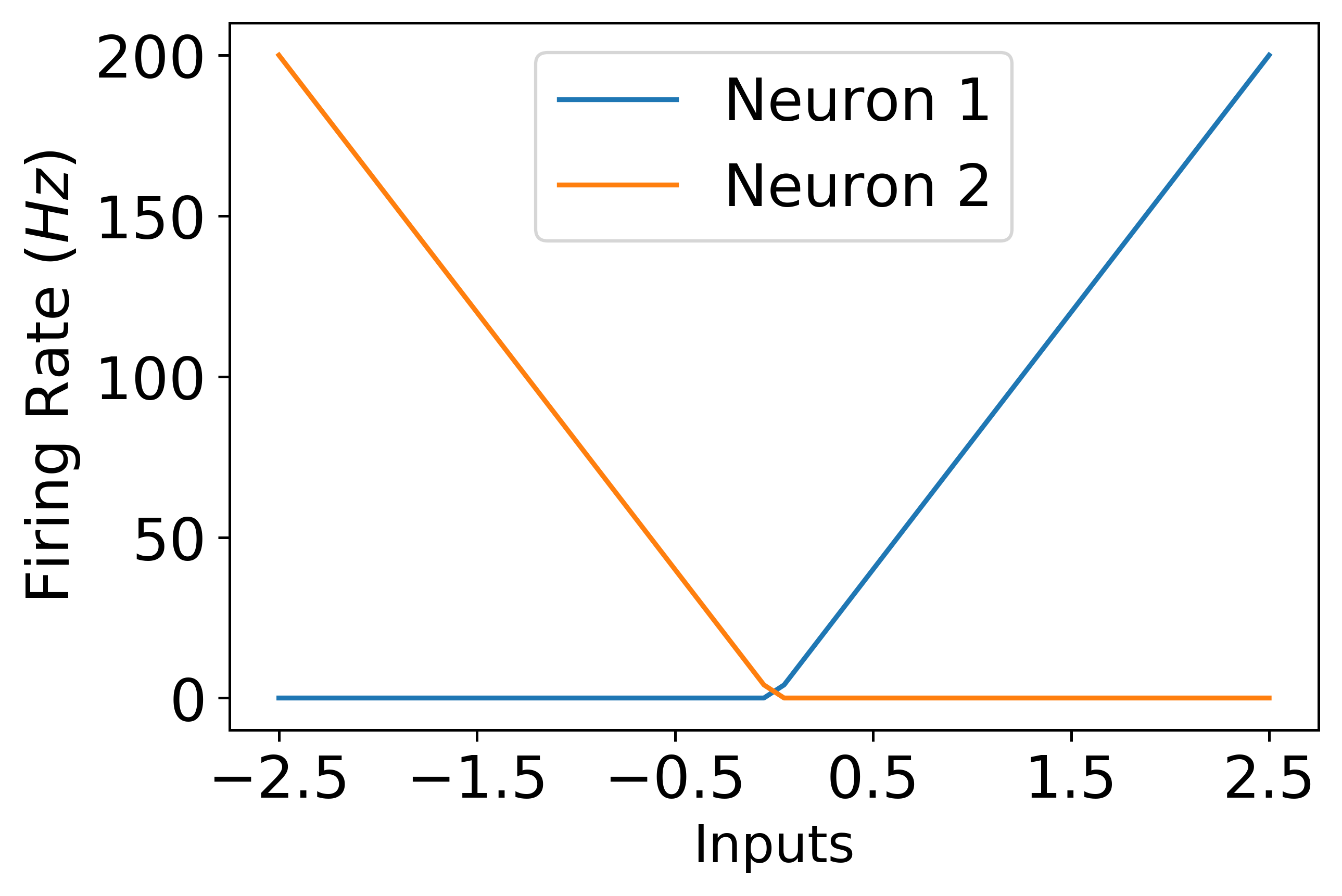}};
\node[] at (-0.5, -1.55) (tc_pic_cap){a. \textit{Tuning Curves}};

\draw[thick, dashed] (0.6, 0) to (0.6, -1.55);
\draw[thick, dashed] (-1.5, 0) to (0.6, 0);

\node[node] at (0.8, 0.5) (A){A};
\node[ens, scale=2.5] at (1.4, -0.3) (B){};
\node[ens_neuron] at (1.4, -0.2) (D){1};
\node[ens_neuron] at (1.4, -0.4) (E){2};
\node[node] at (0.8, -1.1) (C){B};
\node[node] at (2.2, -0.3) (F){O};

\path[->, draw=gray, snake, thick] (F) -> (2.6, -0.3);
\path[->, red] (A) edge [bend right = -15, very thick] node[below=0.1cm, left=0.01cm] {$\frac{1}{2}$} (B.west);
\path[->, green] (A) edge [bend right = -30, very thick] node[above=0.1cm, right=0.2cm] {$\frac{1}{2}$} (F);
\path[->, blue] (C) edge [bend left = -15, very thick] node[above=0.1cm, left=0.01cm] {$\frac{-1}{2}$} (B.west);
\path[->, brown] (C) edge [bend left = -30, very thick] node[below=0.1cm, right=0.2cm] {$\frac{1}{2}$} (F);
\path[->] (D) edge [bend right = -15, very thick] node[above]{$\frac{r}{\phi}$} (F);
\path[->] (E) edge [bend left = -15, very thick] node[below]{$\frac{r}{\phi}$} (F);
\node[draw, rectangle, minimum width=3.8cm, minimum height = 1.4cm] at (-0.35, 0.4) (LEGEND) {};
\matrix [] at (-0.35, 0.4) {
  \node [neuron,label=right:Neuron] {}; &
  \node [node, label=right:Node] {}; \\
  \node [ens, label=right:Ensemble] {}; &
  \node [label=right:Radius] {$r$}; \\
  \node [label=right:MFR] {$\phi$}; \\
};
\node[] at (1.6, -1.55) (tc_pic_cap){b. \textit{$max(a, b)$ Network}};
\end{tikzpicture}
\caption{\textit{Tuning Curves} and \textit{$max(a, b)$ Network}. MFR: Max Firing Rate; Node: A programming construct to either represent a value or sum the inputs, and forward the same.}
\label{fig:avam_max_a_b}
\end{figure}

Therefore, we rather take a unique approach of using the tuning curves to estimate the $|.|$ function. Tuning curves characterize the activation (i.e. firing rate) profile of neurons with respect to the input stimulus. In NEF, these curves depend on the neuron type and properties (e.g. max firing rate $\phi$, representational radius $r$, etc.). Upon configuring an $Ensemble$ of two IF neurons properly (e.g. $\phi = 200$Hz, $r = 2.5$), one can obtain desirable tuning curves as shown in the Fig. \ref{fig:avam_max_a_b}a, which resembles the plot of the $|.|$ function; we leverage the same to estimate the absolute value of a signed scalar input. In our example (Fig. \ref{fig:avam_max_a_b}a), ``Neuron $1$'' (and ``Neuron $2$'') is tuned to fire at $\phi = 200$Hz when $2.5$ (and $-2.5$) is fed to the $Ensemble$. Thus, irrespective of the sign of the input values $\in [-r,\cdots, r]$, we receive a positive firing rate from either neuron. 
Therefore, for an input $r$ (or $-r$), we can filter the spiking output from the corresponding neuron (the other outputs $0$) to obtain $\phi$ and scale it with $\frac{r}{\phi}$ (i.e. $\phi \times \frac{r}{\phi}$) to estimate $|r|$. Note that for such a system of two neurons, one needs to pre-determine a max firing rate $\phi$ and the approximate representational radius $r$. If the magnitude of the input value is lesser (or larger) than $r$, then this system produces a noisier (and saturated) output.

\begin{figure}[h]
\begin{tikzpicture}
\node[] at (0, 0) (mjop_pic) {\includegraphics[scale=0.3]{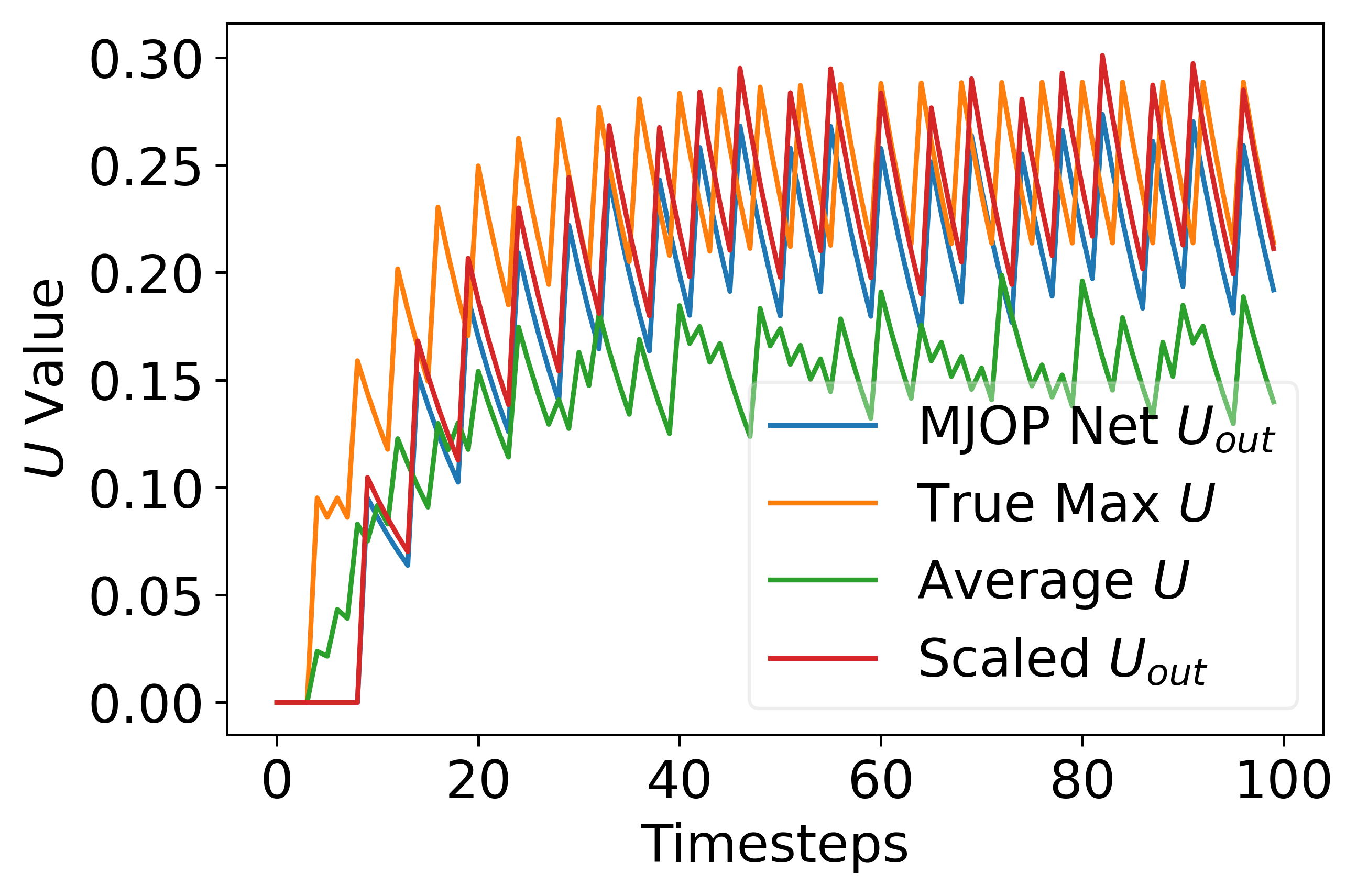}};
\node[] at (4.2, 0) (avam_pic) {\includegraphics[scale=0.3]{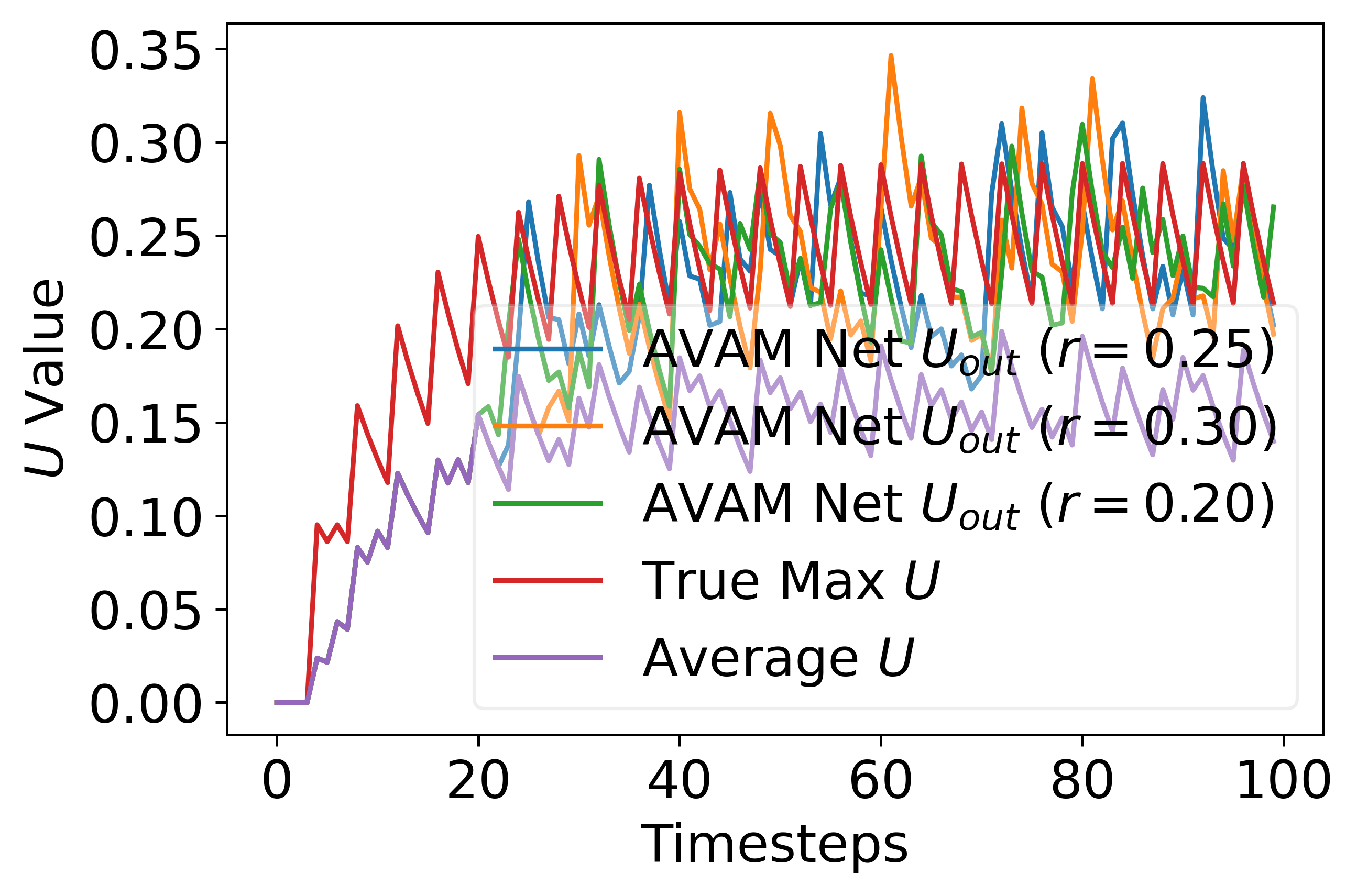}};
\node[] at (0.4, -1.6) (mjop_pic_cap){a. MJOP Net $U_{out}$};
\node[] at (4.5, -1.6) (avam_pic_cap){b. AVAM Net $U_{out}$};
\end{tikzpicture}
\caption{``MJOP Net $U_{out}$'' is the synapsed/filtered spiking output from the soma. It is scaled by $1.1$ (to obtain ``Scaled $U_{out}$'') to match the ``True Max $U$'' $=max(U_1, U_2, U_3, U_4)$ closely. ``AVAM Net $U_{out}$'' are the outputs from Node \texttt{O} (in Fig. \ref{fig:avam_max_poc}) for different radii $r$. For $r = 0.25$ and $0.20$, the $U_{out}$ matches the ``True Max $U$'' $=max(U_1, U_2, U_3, U_4)$ closely.}
\label{fig:max_jop_poc}
\end{figure}

\subsubsection{Estimation of $max(a, b)$: }The above method of estimating the $|.|$ function can be incorporated with the average term calculation to construct a network as shown in the Fig. \ref{fig:avam_max_a_b}b to estimate the $max(a, b)$. In Fig. \ref{fig:avam_max_a_b}b, Nodes \texttt{A} and \texttt{B} represent and output the values $a$ and $b$ respectively as currents. They are directly connected to the Node \texttt{O} with a connection weight of $1/2$ each. Thus, their scaled output i.e. $a/2$ and $b/2$ gets summed up at Node \texttt{O} to result in $(a+b)/2$. Nodes \texttt{A} and \texttt{B} are also connected to an $Ensemble$ of two Neurons, \texttt{1} and \texttt{2} (whose tuning curves are similar to that in Fig. \ref{fig:avam_max_a_b}a and their representational radius $r \approx (|a-b|)/2$) with a connection weight of $1/2$ and $-1/2$ respectively, such that the sum $(a - b)/2$ fed to the $Ensemble$. Note that $r$ can be heuristically set without knowing the actual values of $a$ and $b$ (shown later). Depending on the sign of the sum $(a - b)/2$, either the Neuron \texttt{1} or Neuron \texttt{2} spikes at a frequency $\phi$ (the other outputs $0$). Therefore, after filtering the spike outputs to obtain $\phi$ and scaling it with $\frac{r}{\phi}$ ($= \phi \times \frac{r}{\phi}$) through the weighted connection to the Node \texttt{O}, $r$ is sent. The Node \texttt{O} then finally accumulates the inputs, i.e. $\frac{a+b}{2} + r \approx \frac{a+b}{2} + \frac{|a-b|}{2}$ which is approximately equal to the $max(a, b)$ and relays the same.

\begin{figure}
\begin{tikzpicture}[
scale=2, 
neuron/.style={draw, circle, fill=yellow},
ens_neuron/.style={draw, circle, fill=yellow, scale=0.5},
ens/.style={draw, regular polygon, regular polygon sides=6, scale=1, top color =white , bottom color = orange!20},
node/.style={draw, circle, fill=violet!40},
array/.style={matrix of nodes, nodes={draw, minimum size=5mm, fill=red!10},column sep=-\pgflinewidth, row sep=0.5mm, nodes in empty cells},
]
\matrix[array] at (0, -1.25) (pw) {1&2\\
                                4&3\\};

\node[node] at (1.0, -0.3) (1){1};
\node[ens, scale=1.5] at (1.5, -0.6) (2){};
\node[ens_neuron] at (1.5, -0.55) (3){};
\node[ens_neuron] at (1.5, -0.65) (4){};
\node[node] at (1.0, -0.9) (5){2};
\node[node] at (2.25, -0.6) (6){P};

\path[->, red] (1) edge [bend right = -15, very thick] node[below=0.1cm, left=0.01cm] {} (2.west);
\path[->, green] (1) edge [bend right = -30, very thick] node[above=0.1cm, right=0.2cm] {} (6);
\path[->, blue] (5) edge [bend left = -15, very thick] node[above=0.1cm, left=0.01cm] {} (2.west);
\path[->, brown] (5) edge [bend left = -30, very thick] node[below=0.1cm, right=0.2cm] {} (6);
\path[->] (3) edge [bend right = -15, very thick] node[above]{} (6);
\path[->] (4) edge [bend left = -15, very thick] node[below]{} (6);


\node[node] at (1.0, -1.6) (7){3};
\node[ens, scale=1.5] at (1.5, -1.9) (8){};
\node[ens_neuron] at (1.5, -1.85) (9){};
\node[ens_neuron] at (1.5, -1.95) (10){};
\node[node] at (1.0, -2.2) (11){4};
\node[node] at (2.25, -1.9) (12){Q};

\path[->, red] (7) edge [bend right = -15, very thick] node[below=0.1cm, left=0.01cm] {} (8.west);
\path[->, green] (7) edge [bend right = -30, very thick] node[above=0.1cm, right=0.2cm] {} (12);
\path[->, blue] (11) edge [bend left = -15, very thick] node[above=0.1cm, left=0.01cm] {} (8.west);
\path[->, brown] (11) edge [bend left = -30, very thick] node[below=0.1cm, right=0.2cm] {} (12);
\path[->] (9) edge [bend right = -15, very thick] node[above]{} (12);
\path[->] (10) edge [bend left = -15, very thick] node[below]{} (12);

\node[ens, scale=1.5] at (2.75, -1.25) (13){};
\node[ens_neuron] at (2.75, -1.20) (14){};
\node[ens_neuron] at (2.75, -1.30) (15){};
\node[node] at (3.5, -1.25) (16){O};

\path[->, red] (6) edge [bend right = -15, very thick] node[below=0.1cm, left=0.01cm] {} (13.west);
\path[->, blue] (12) edge [bend left = -15, very thick] node[above=0.1cm, left=0.01cm] {} (13.west);
\path[->, green] (6) edge [bend right = -30, very thick] node[above=0.1cm, right=0.2cm] {} (16);
\path[->, brown] (12) edge [bend left = -30, very thick] node[below=0.1cm, right=0.2cm] {} (16);
\path[->] (14) edge [bend right = -15, very thick] node[above]{} (16);
\path[->] (15) edge [bend left = -15, very thick] node[below]{} (16);

\draw[dotted, very thick] (pw-1-1.north) to (-0.125, -0.3);
\draw[dotted, very thick] (pw-1-2.north) to (0.125, -0.9);
\draw[dotted, very thick] (pw-2-2.south) to (0.125, -1.6);
\draw[dotted, very thick] (pw-2-1.south) to (-0.125, -2.2);

\draw[dotted, very thick] (-0.125, -0.3) to (0.5, -0.3);
\draw[dotted, very thick] (0.125, -0.9) to (0.5, -0.9);
\draw[dotted, very thick] (0.125, -1.6) to (0.5, -1.6);
\draw[dotted, very thick] (-0.125, -2.2) to (0.5, -2.2);

\path[->, draw=gray, snake, thick] (16) -> (4.0, -1.25);
\path[->, draw=cyan, snake, thick] (0.5, -0.3) -> (1);
\path[->, draw=cyan, snake, thick] (0.5, -0.9) -> (5);
\path[->, draw=cyan, snake, thick] (0.5, -1.6) -> (7);
\path[->, draw=cyan, snake, thick] (0.5, -2.2) -> (11);

\draw[thick] (0.4, -0.3) -> (0.4, -0.1);
\draw[thick] (0.2, -0.3) -> (0.2, -0.1);
\draw[thick] (0.0, -0.3) -> (0.0, -0.1);

\draw[thick] (0.3, -0.9) -> (0.3, -0.7);
\draw[thick] (0.15, -0.9) -> (0.15, -0.7);

\draw[thick] (0.4, -1.6) -> (0.4, -1.4);

\draw[thick] (0.05, -2.2) -> (0.05, -2.0);
\draw[thick] (0.15, -2.2) -> (0.15, -2.0);
\draw[thick] (0.25, -2.2) -> (0.25, -2.0);
\draw[thick] (0.35, -2.2) -> (0.35, -2.0);
\node[draw, rectangle, minimum width=1.25cm, minimum height = 0.5cm] at (3.55, -2.0) (LEGEND) {};
\matrix [] at (3.55, -2.0) {
  \node [draw, scale=1.7, fill=red!10,label=right:MP]{};\\
};
\end{tikzpicture}
\caption{\textit{AVAM Net for $2\times2$ MaxPooling}. MP: $2\times2$ MaxPooling Window. Edges are color coded; refer Fig. \ref{fig:avam_max_a_b} legend.}
\label{fig:avam_max_poc}
\end{figure}
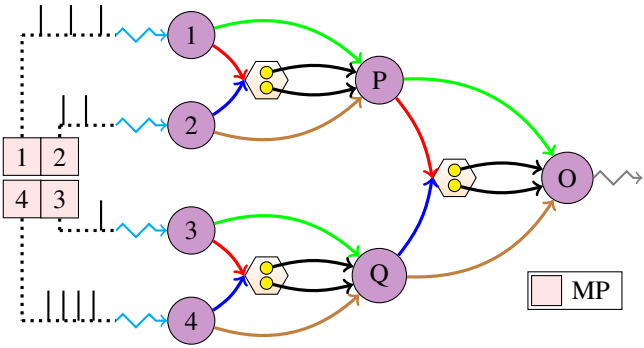

\subsubsection{Proof-of-Concept Demonstration: } For a $2\times2$ MaxPooling window, we can compute the $max(a_1, a_2, a_3, a_4)$ as $max(max(a_1, a_2), max(a_3, a_4))$ using the Associative Property of $max()$. We therefore construct an example AVAM Net, shown in Fig. \ref{fig:avam_max_poc} with four spiking inputs (firing time-period $= 10, 8, 4, 6$) connected to the individual Nodes. Note that the filtered spikes i.e. currents $U_i$ are being fed to the Nodes (here $a_i = U_i$) and they relay the same to the next connected components. In this hierarchical network, $max(a_1, a_2)$ and $max(a_3, a_4)$ gets estimated at the Nodes \texttt{P} and \texttt{Q} respectively. They forward the same and Node \texttt{O} finally estimates the $max(max(a_1, a_2), max(a_3, a_4))$. Three instances of this AVAM Net were executed on the Loihi chip with IF neurons' $\phi$ fixed to $500$Hz and $r \in$ \{$0.20, 0.25, 0.30$\}. In each instance, all the neurons in each $Ensemble$ had the same $\phi$ and $r$. Each instance's estimated $max(U_1, U_2, U_3, U_4)$ i.e. ``AVAM Net $U_{out}$'' (for a corresponding $r$) is shown in the Fig. \ref{fig:max_jop_poc}b.  We see that $r$'s value around $0.25$ (for a fixed $\phi$) does a fair job of approximating the ``True Max U''; as well as, the estimated max $U$ are higher than the Average $U$.
\\~\\
\noindent One should note that ideally, the output current from a MaxPooling window (of spiking neurons) should be the current due to the maximally firing neuron (i.e. ``Exact Max $U$''). However, the MJOP and AVAM spiking-MaxPooling methods compute the instantaneous max of all incoming currents (in a pooling window) at each time-step, which is not equivalent to the ``Exact Max $U$''. It is possible that the instantaneous max of currents could be higher than the ``exact max'' when a slower spiking neuron fires recently than the maximally firing neuron. This holds true with ``True Max $U$'' as well; however, we defined it as $max()$ of currents in the spirit of MaxPooling in ANNs. Therefore, our methods are approximations of the ideal MaxPooling in SNNs.

\subsection{Heuristics for $scale$ (MJOP) and $radius$ (AVAM)}
As seen in the \textbf{Proof-of-Concept Demonstration} sections, one needs to properly $scale$ the $U_{out}$ in case of MJOP Net and set the $radius$ parameter in AVAM Net to correctly approximate the ``True Max $U$''. For a fixed configuration of the compartments/neurons in the MJOP and AVAM Net, the value of $scale$ and $radius$ depends on the group of inputs $U_i$. Recollect that in the MJOP Net, the soma compartment spikes at a rate \textit{corresponding} to the maximum input $U_i$, and the output $U_{out}$ needs to be scaled accordingly. And in the AVAM Net, the $radius$ should be heuristically chosen to be equal to $|a-b|/2$ for estimating the $max(a, b)$ (where $a$ and $b$ are the inputs $U_i$). The inputs $U_i$ in turn depend on the periodicity (or the Inter-Spike Interval (ISI)) of the incoming individual spike trains. We note here that the maximum and minimum value of $U_i$ can be $1$ (with spike amplitude $=1$, ISI $=1$) and $0$ (with the corresponding neuron not spiking at all) respectively. This implies that the maximum value of the difference of two $U_i$s can be $1$, i.e. $radius \leq 1$ always. This holds true for the $radius$ value of the $Ensemble$ neurons further in the AVAM Net hierarchy. Moreover, many or all of the neurons in a pooling window may spike with ISI $>1$, which would further lower down the $radius$ value.

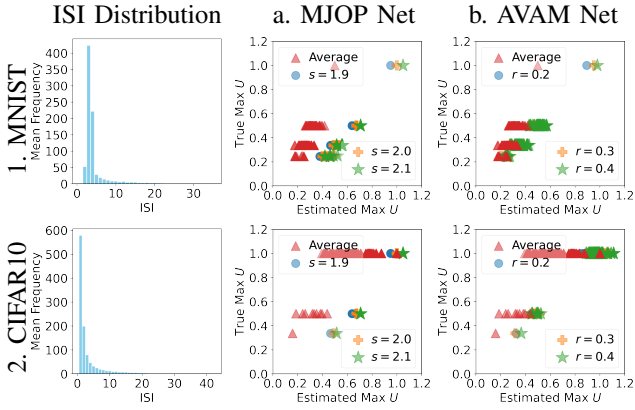
\begin{figure}[h]
\begin{tikzpicture}
\node[] at (0.2, 0) (A) {ISI Distribution};
\node[] at (2.95, 0) (B) {a. MJOP Net};
\node[] at (5.6, 0) (C) {b. AVAM Net};

\node[rotate=90] at (-1.4, -1.4) (D) {1. MNIST};
\node[rotate=90] at (-1.4, -3.9) (D) {2. CIFAR10};

\node[] at (0.05, -1.5) (isi_mnist) {\includegraphics[scale=0.25]{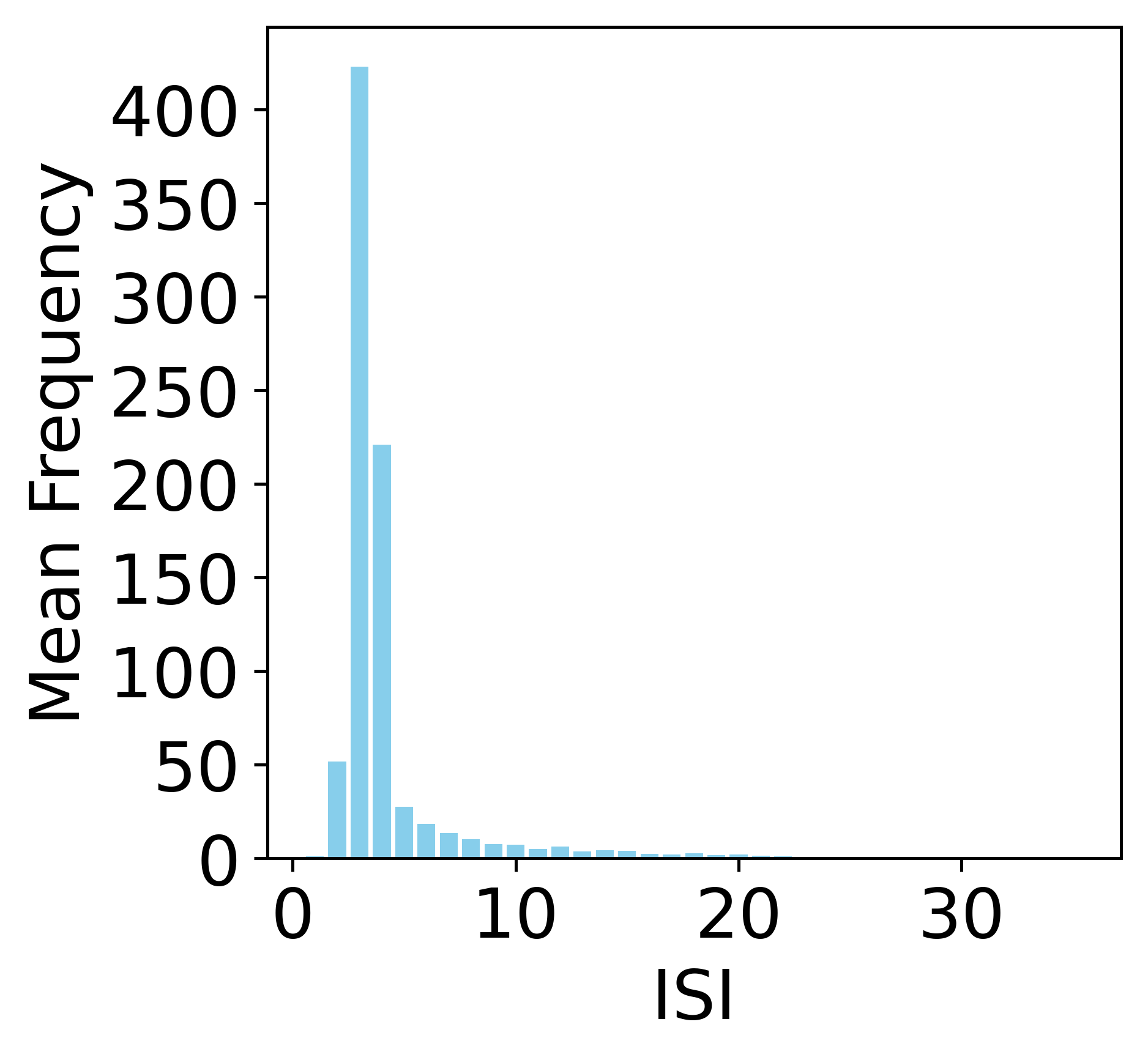}};
\node[] at (0.05, -4.0) (isi_cifar10)
{\includegraphics[scale=0.25]{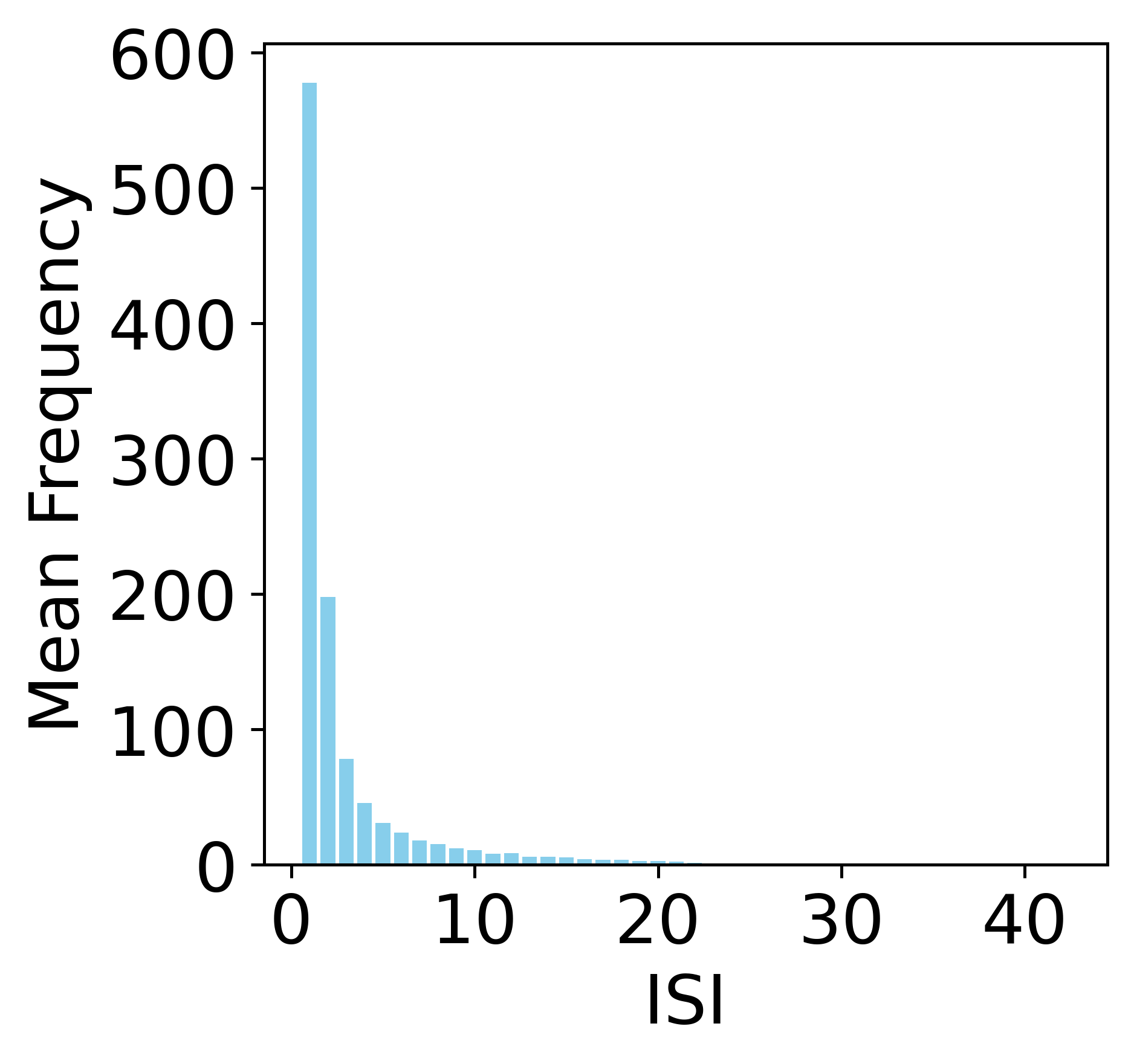}};

\node[] at (2.8, -1.5) (mjop_mnist) {
\includegraphics[scale=0.25]{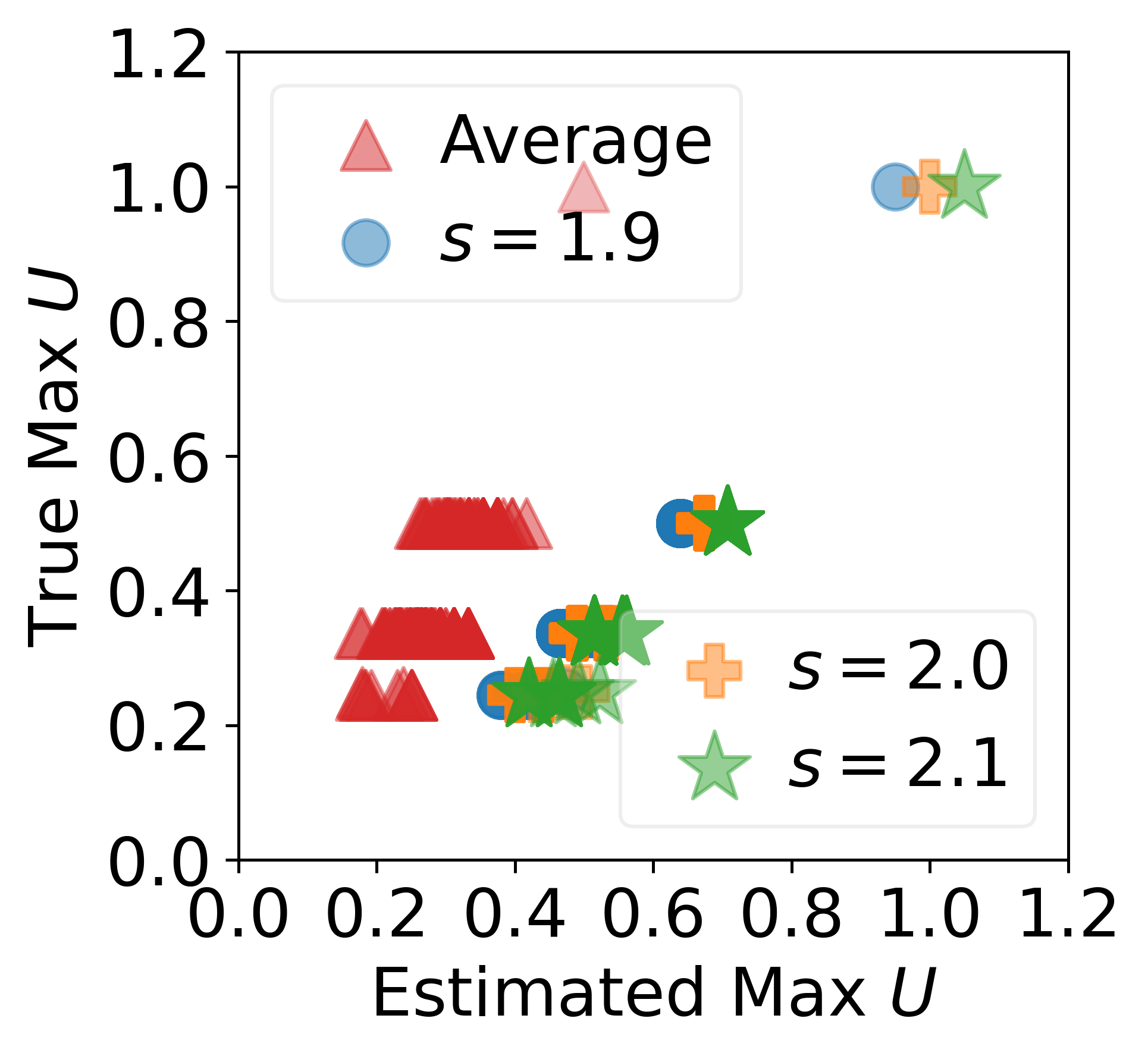}
};
\node[] at (2.8, -4) (mjop_cifar10) {
\includegraphics[scale=0.25]{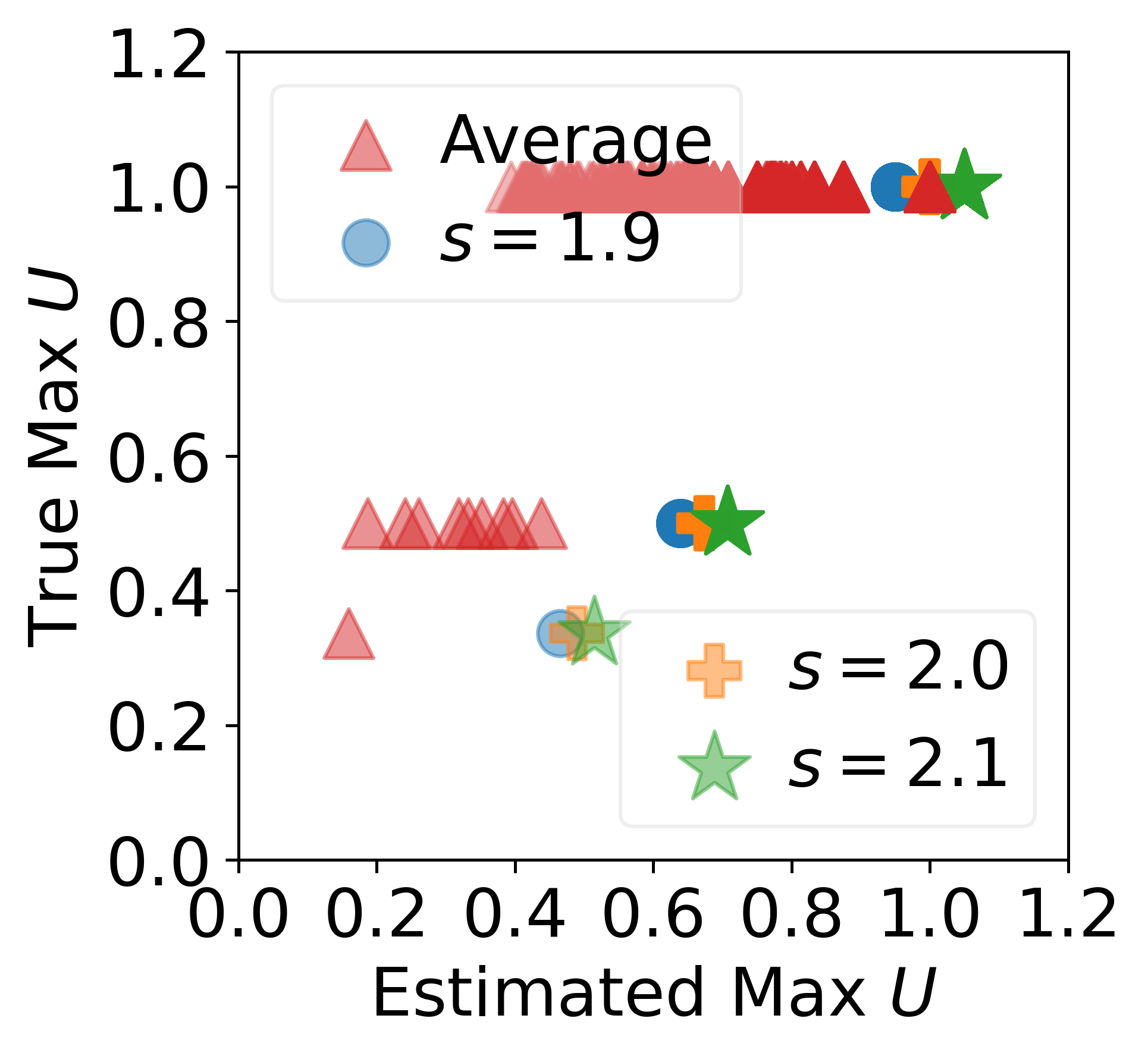}
};

\node[] at (5.5, -1.5) (avam_mnist) {
\includegraphics[scale=0.25]{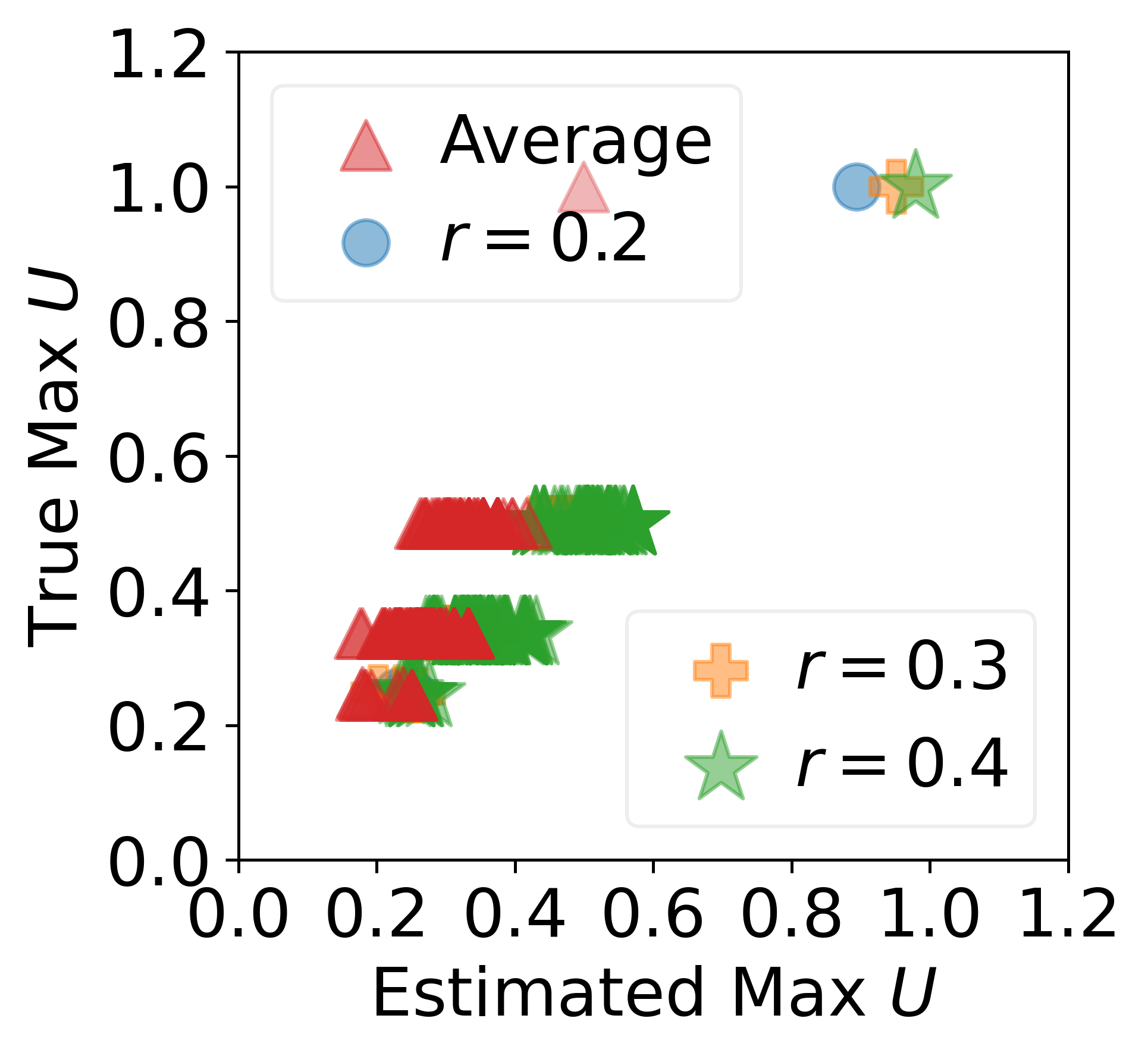}
};
\node[] at (5.5, -4) (avam_cifar10) {
\includegraphics[scale=0.25]{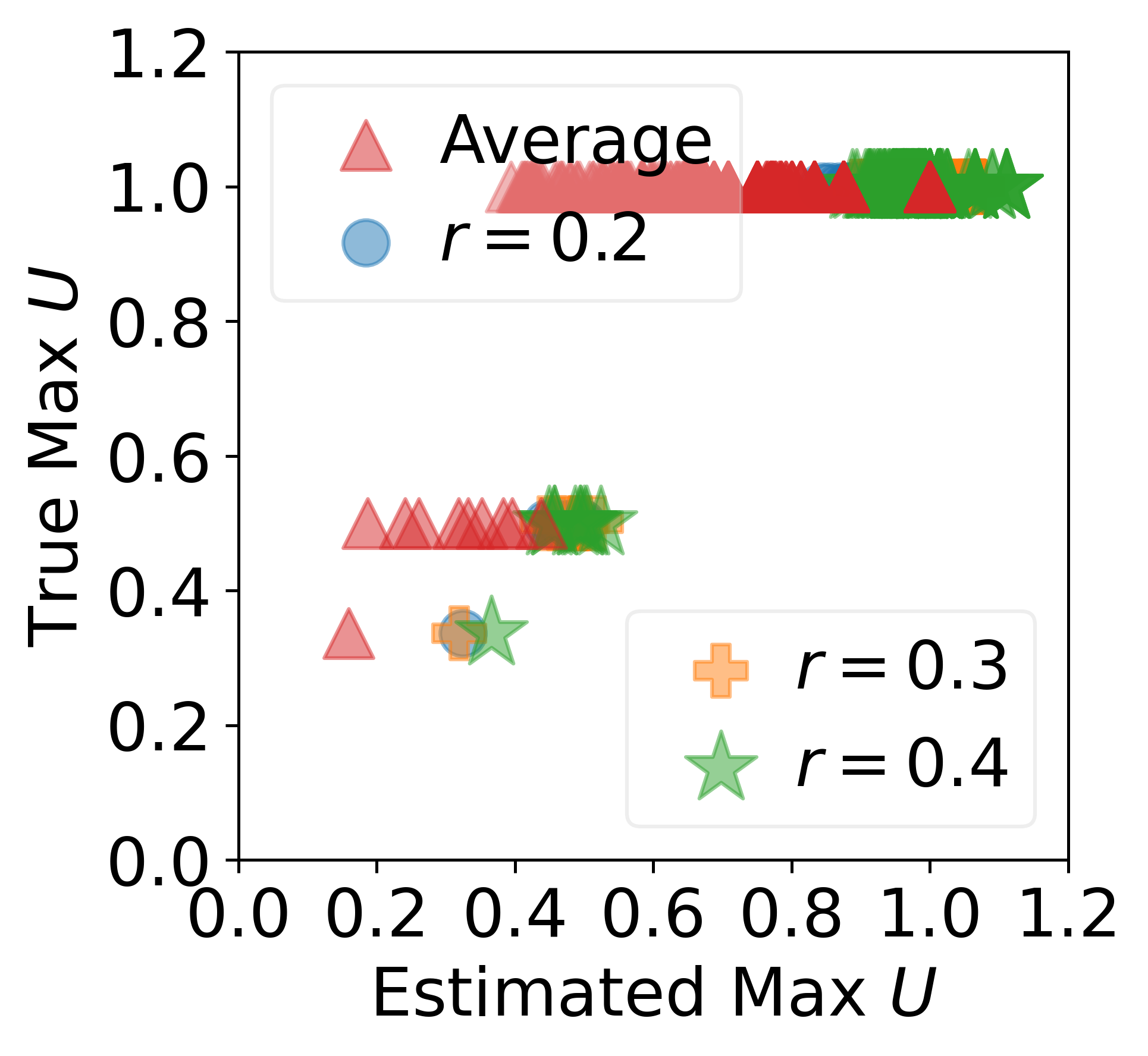}
};
\end{tikzpicture}
\caption{\textit{Scatter-Plots of MJOP \& AVAM Net Outputs} over spiking inputs (their ISI's in groups of $4$). $s$: Scale, $r$: Radius. Corr. Coef.: ($1$, a) $0.98$; ($1$, b) $0.99$; ($2$, a) $0.99$, ($2$, b) $0.95$.}
\label{fig:tuning_scale_radius}
\end{figure}

One can heuristically choose the $scale$ and $radius$ values by analysing the ISI distribution of the neurons in a model's Conv layer (preceding the MaxPooling layer) for a dataset; although, these values may be required to be tuned further. We therefore conduct a toy experiment where we collect the ISIs of $2048$ neurons in a Conv layer (of one of our converted SNNs) for $1000$ training images of MNIST and CIFAR10 each, and construct the respective ISI distributions (averaged across $1000$ images). From each distribution, we obtain $256$ groups of randomly sampled ISIs (group size $4$) and filter the respective spiking inputs to create $256$ groups of \{$U_1, U_2, U_3, U_4$\}. We next compute the ``True Max $U$'' and ``Estimated Max $U$'' for different values of $scale$ and $radius$ in the MJOP \& AVAM Net respectively, and analyse their effects (refer Fig. \ref{fig:tuning_scale_radius}). Note that the ``Average'' $U$ values (synonymous to AveragePooling) are $\leq$ the respective estimated max $U$. The high correlation of the true and estimated $U$ obtained for a varied group of spiking inputs shows the efficacy of our spiking-MaxPooling methods. For our MJOP \& AVAM Net, $scale$ value around $2.0$ and a $radius$ value around $0.3$ fairly approximates the ``True Max $U$''. Note that in the AVAM Net, ideally, the $radius$ of each $Ensemble$'s neurons should be set independently to estimate the varying $|a-b|/2$, however for simplicity, we keep it same.

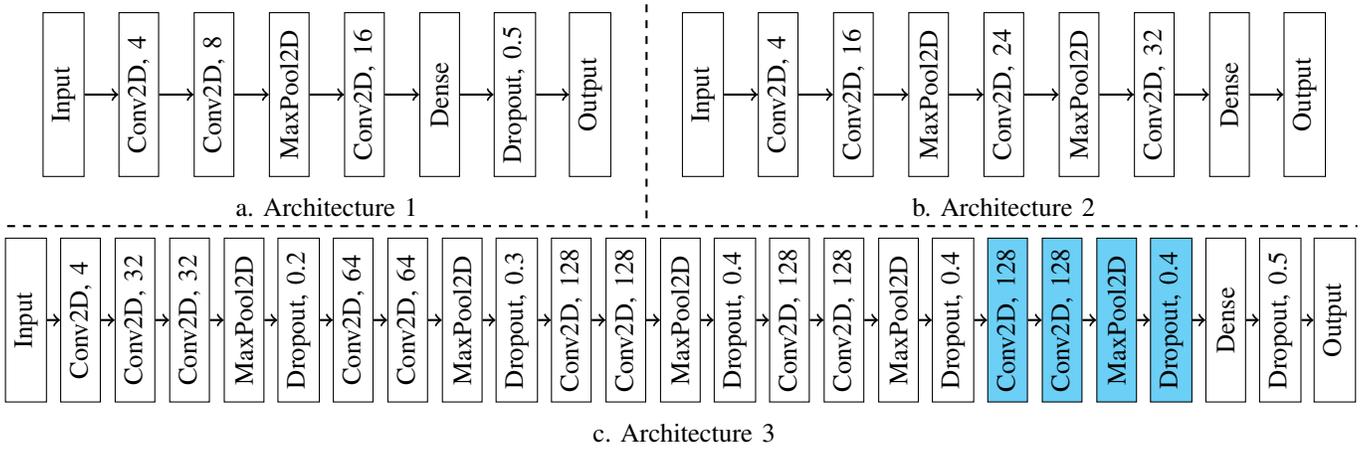
\begin{figure*}
\begin{tikzpicture}[block/.style={draw, rectangle,minimum height=1.5em,minimum width=6.2em, rotate=90}]


\node[block] at (0.25, 0) (A){Input};
\node[block] at (1.25, 0) (B){Conv2D, 4};
\node[block] at (2.25, 0) (C){Conv2D, 8};
\node[block] at (3.25, 0) (D){MaxPool2D};
\node[block] at (4.25, 0) (E){Conv2D, 16};
\node[block] at (5.25, 0) (F){Dense};
\node[block] at (6.25, 0) (F1){Dropout, 0.5};
\node[block] at (7.25, 0) (G){Output};

\draw[thick, ->] (A) to (B);
\draw[thick, ->] (B) to (C);
\draw[thick, ->] (C) to (D);
\draw[thick, ->] (D) to (E);
\draw[thick, ->] (E) to (F);
\draw[thick, ->] (F) to (F1);
\draw[thick, ->] (F1) to (G);

\node[] at (3.75, -1.5) (Arch1){a. Architecture 1};

\draw[thick, dashed] (8, 1.2) to (8, -1.75);
\draw[thick, dashed] (-0.5, -1.75) to (17.45, -1.75);

\node[block] at (8.75, 0) (I){Input};
\node[block] at (9.75, 0) (J){Conv2D, 4};
\node[block] at (10.75, 0) (K){Conv2D, 16};
\node[block] at (11.75, 0) (L){MaxPool2D};
\node[block] at (12.75, 0) (M){Conv2D, 24};
\node[block] at (13.75, 0) (N){MaxPool2D};
\node[block] at (14.75, 0) (O){Conv2D, 32};
\node[block] at (15.75, 0) (P){Dense};
\node[block] at (16.75, 0) (Q){Output};

\draw[thick, ->] (I) to (J);
\draw[thick, ->] (J) to (K);
\draw[thick, ->] (K) to (L);
\draw[thick, ->] (L) to (M);
\draw[thick, ->] (M) to (N);
\draw[thick, ->] (N) to (O);
\draw[thick, ->] (O) to (P);
\draw[thick, ->] (P) to (Q);

\node[] at (12.75, -1.5) (Arch2){b. Architecture 2};

\node[block] at (-0.25, -3) (R){Input};
\node[block] at (0.475, -3) (S){Conv2D, 4};

\node[block] at (1.2, -3) (T){Conv2D, 32};
\node[block] at (1.925, -3) (U){Conv2D, 32};
\node[block] at (2.65, -3) (V){MaxPool2D};
\node[block] at (3.375, -3) (W){Dropout, 0.2};

\node[block] at (4.1, -3) (X){Conv2D, 64};
\node[block] at (4.825, -3) (Y){Conv2D, 64};
\node[block] at (5.55, -3) (Z){MaxPool2D};
\node[block] at (6.275, -3) (Z1){Dropout, 0.3};

\node[block] at (7.0, -3) (Z2){Conv2D, 128};
\node[block] at (7.725, -3) (Z3){Conv2D, 128};
\node[block] at (8.45, -3) (Z4){MaxPool2D};
\node[block] at (9.175, -3) (Z5){Dropout, 0.4};

\node[block] at (9.9, -3) (Z6){Conv2D, 128};
\node[block] at (10.625, -3) (Z7){Conv2D, 128};
\node[block] at (11.35, -3) (Z8){MaxPool2D};
\node[block] at (12.075, -3) (Z9){Dropout, 0.4};

\node[fill=cyan!50, block] at (12.8, -3) (A1){Conv2D, 128};
\node[fill=cyan!50, block] at (13.525, -3) (A2){Conv2D, 128};
\node[fill=cyan!50, block] at (14.25, -3) (A3){MaxPool2D};
\node[fill=cyan!50, block] at (14.975, -3) (A4){Dropout, 0.4};

\node[block] at (15.7, -3) (A5){Dense};
\node[block] at (16.425, -3) (A6){Dropout, 0.5};

\node[block] at (17.15, -3) (A7){Output};

\draw[thick, ->] (R) to (S);
\draw[thick, ->] (S) to (T);
\draw[thick, ->] (T) to (U);
\draw[thick, ->] (U) to (V);
\draw[thick, ->] (V) to (W);
\draw[thick, ->] (W) to (X);
\draw[thick, ->] (X) to (Y);
\draw[thick, ->] (Y) to (Z);
\draw[thick, ->] (Z) to (Z1);
\draw[thick, ->] (Z1) to (Z2);
\draw[thick, ->] (Z2) to (Z3);
\draw[thick, ->] (Z3) to (Z4);
\draw[thick, ->] (Z4) to (Z5);
\draw[thick, ->] (Z5) to (Z6);
\draw[thick, ->] (Z6) to (Z7);
\draw[thick, ->] (Z7) to (Z8);
\draw[thick, ->] (Z8) to (Z9);
\draw[thick, ->] (Z9) to (A1);
\draw[thick, ->] (A1) to (A2);
\draw[thick, ->] (A2) to (A3);
\draw[thick, ->] (A3) to (A4);
\draw[thick, ->] (A4) to (A5);
\draw[thick, ->] (A5) to (A6);
\draw[thick, ->] (A6) to (A7);

\node[] at (8.5, -4.5) (Arch3){c. Architecture 3};

\end{tikzpicture}
\caption{\textit{CNN Architectures.} ``Conv2D, x'' \hspace{0.05pt} $\Rightarrow$ \hspace{0.05pt} ``x'' number of filters in the Conv layer; ``Dropout, x'' \hspace{0.05pt} $\Rightarrow$ \hspace{0.05pt} ``x'' dropout probability. In each architecture, Conv layer strides $=(1, 1)$, kernel size $=(3, 3)$ (except the first Conv layer with kernel size $=(1, 1)$); $2 \times 2$ MaxPooling window; $128$ neurons in Dense layer; no activation in the Output layer; no $bias$ in all layers except Dense. MaxPool layers in all architectures have no padding; Conv layers too have no padding except in Architecture $3$. For experiments with MNIST \& FMNIST, the colored blocks in Architecture 3 are removed to account for their smaller image size than CIFAR10. During the inference phase with the SNNs obtained after the conversion of ANNs, the ``Dropout, x'' layers are removed.}
\label{fig:archs}
\end{figure*}
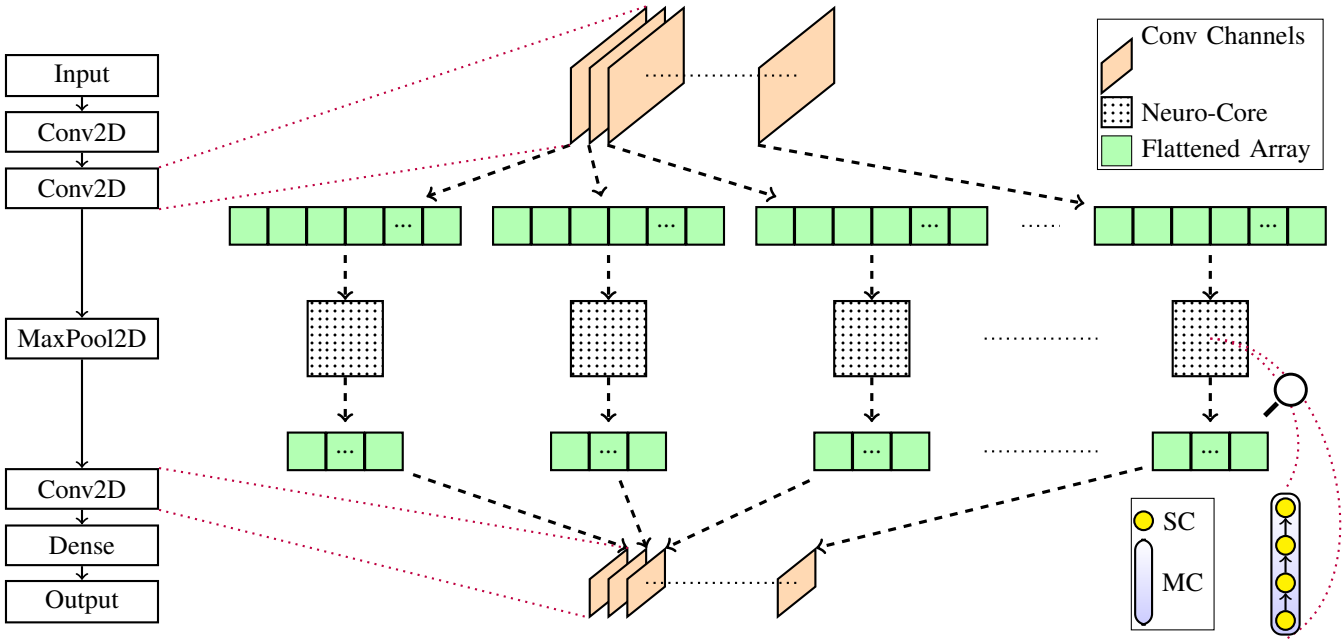
\begin{figure*}
\begin{tikzpicture}[
block/.style={draw, rectangle,minimum height=1.5em,minimum width=5.75em, thick},
channel/.style={draw, rectangle, minimum width=1cm, minimum height = 1cm, thick, fill=orange!30},
core/.style={draw, rectangle, minimum width=1cm, minimum height = 1cm, thick, pattern=dots}, array/.style={matrix of nodes, nodes={draw, minimum size=5mm, fill=green!30, anchor=center, thick},column sep=-\pgflinewidth, row sep=0.5mm, nodes in empty cells},
neuron/.style={draw, circle, fill=yellow, scale=0.75, thick},
multi_compartment/.style={draw, rectangle, minimum width=1.5cm, minimum height = 0.3cm, rounded corners, top color = white, bottom color = blue!20, thick},
]

\node[block] at (0, 0) (A) {Input};
\node[block] at (0, -0.75) (B) {Conv2D};
\node[block] at (0, -1.5) (C) {Conv2D};
\node[block] at (0, -3.5) (D) {MaxPool2D};
\node[block] at (0, -5.5) (E) {Conv2D};
\node[block] at (0, -6.25) (F) {Dense};
\node[block] at (0, -7) (G) {Output};

\draw[thick, ->] (A) to (B);
\draw[thick, ->] (B) to (C);
\draw[thick, ->] (C) to (D);
\draw[thick, ->] (D) to (E);
\draw[thick, ->] (E) to (F);
\draw[thick, ->] (F) to (G);
\node[yslant=0.8, channel] at (7, -0) (1) {};
\node[yslant=0.8, channel] at (7.25, -0) (2) {};
\node[yslant=0.8, channel] at (7.5, -0) (3) {};
\node[yslant=0.8, channel] at (9.5, -0) (5) {};
\path[-] (7.5, -0) edge [dotted, thick] (9.5, -0);


\matrix[array] at (3.5, -2) (H) {& & & &...&\\};
\matrix[array] at (7, -2) (I) {& & & &...&\\};
\matrix[array] at (10.5, -2) (J) {& & & &...&\\};
\matrix[array] at (15, -2) (K) {& & & &...&\\};

\draw[thick, dotted] (12.5, -2) to (13, -2);

\draw[very thick, dashed, ->] (1.south west) to (H);
\draw[very thick, dashed, ->] (2.south west) to (I);
\draw[very thick, dashed, ->] (3.south west) to (J);
\draw[very thick, dashed, ->] (5.south west) to (K);

\node[core] at (3.5, -3.5) (L) {};
\node[core] at (7, -3.5) (M) {};
\node[core] at (10.5, -3.5) (N) {};
\node[core] at (15, -3.5) (O) {};

\path[-] (12, -3.5) edge [dotted, thick] (13.5, -3.5);

\draw[very thick, dashed, ->] (H) to (L);
\draw[very thick, dashed, ->] (I) to (M);
\draw[very thick, dashed, ->] (J) to (N);
\draw[very thick, dashed, ->] (K) to (O);

\matrix[array] at (3.5, -5) (P) {&...&\\};
\matrix[array] at (7, -5) (Q) {&...&\\};
\matrix[array] at (10.5, -5) (R) {&...&\\};
\matrix[array] at (15, -5) (S) {&...&\\};

\draw[thick, dotted] (12, -5) to (13.5, -5);

\draw[very thick, dashed, ->] (L) to (P);
\draw[very thick, dashed, ->] (M) to (Q);
\draw[very thick, dashed, ->] (N) to (R);
\draw[very thick, dashed, ->] (O) to (S);

\node[yslant=0.8, channel, scale=0.5] at (7, -6.75) (6) {};
\node[yslant=0.8, channel, scale=0.5] at (7.25, -6.75) (7) {};
\node[yslant=0.8, channel, scale=0.5] at (7.5, -6.75) (8) {};
\node[yslant=0.8, channel, scale=0.5] at (9.5, -6.75) (9) {};


\draw[very thick, dashed, ->] (P) to (6.north east);
\draw[very thick, dashed, ->] (Q) to (7.north east);
\draw[very thick, dashed, ->] (R) to (8.north east);
\draw[very thick, dashed, ->] (S) to (9.north east);
\path[-] (7.5, -6.75) edge [dotted, thick] (9.5, -6.75);

\draw[thick, dotted, purple] (C.north east) to (1.north east);
\draw[thick, dotted, purple] (C.south east) to (1.south west);

\draw[thick, dotted, purple] (E.north east) to (6.north east);
\draw[thick, dotted, purple] (E.south east) to (6.south west);
\draw[thick, dotted, bend right = -60, purple] (15, -3.5) to (16, -5.50);
\draw[thick, dotted, bend right = -75, purple] (15, -3.5) to (16, -7.50);
\node[scale=1] at (16, -4.25) (F1){\bcloupe};

\node[multi_compartment, scale=1.25, rotate=90] at (16, -6.50) (A1){};
\node[neuron] at (16, -5.75) (B1){};
\node[neuron] at (16, -6.25) (C1){};
\node[neuron] at (16, -6.75) (D1){};
\node[neuron] at (16, -7.25) (E1){};

\draw[thick, ->] (C1) to (B1);
\draw[thick, ->] (D1) to (C1);
\draw[thick, ->] (E1) to (D1);

\node[draw, rectangle, minimum width=3cm, minimum height = 2cm] at (15.00, -0.25) (LEGEND1) {};
\matrix [] at (15.00, -0.25) {
  \node[scale=0.4, yslant=0.8, channel, label=right:Conv Channels]{};\\
  \node[scale=0.4, core, label=right:Neuro-Core]{};\\
  \node[draw, scale=1.7, fill=green!30, label=right:Flattened Array]{};\\
};

\node[draw, rectangle, minimum width=1.1cm, minimum height = 1.75cm] at (14.5, -6.50) (LEGEND2) {};
\matrix [] at (14.5, -6.50) {
  \node [neuron, label=right:SC]{}; \\
  \node [multi_compartment, scale=0.75, rotate=90,
  label=below:MC]{}; \\
};
\end{tikzpicture}
\caption{\textit{MAX join-Op based MaxPooling.} For a MaxPool layer in an architecture, the preceding Conv layer's channels are each flattened and mapped to a single Neuro-Core. Each Neuro-Core has an $Ensemble$ of MJOP Net configured MC neurons. Post execution of MJOP Nets on the Neuro-Cores, the flattened vectors are reshaped to channels and passed to the next Conv layer.}
\label{fig:max_jop_mp}
\end{figure*}

\begin{table*}[]
\fontsize{8.5pt}{8.5pt}\selectfont
\centering
\setlength\tabcolsep{0.6pt}
\begin{tabular}{|c|c|c|c|c|c|c|c|c|c|c|c|c|c|c|c|c|c|c|c|c|c|c|c|c|c|c|c|}
\hline
\multirow{3}{*}{} & \multicolumn{10}{c|}{\normalsize \textbf{Architecture 1}}                                                                                       & \multicolumn{10}{c|}{\normalsize \textbf{Architecture 2}}                                                                                       & \multicolumn{7}{c|}{\normalsize \textbf{Architecture 3}}                                                            \\ \cline{2-28}
                  & \multirow{2}{*}{\textbf{NSR}} & \multirow{2}{*}{\textbf{TMS}} & \multirow{2}{*}{\textbf{MAS}} & \multicolumn{3}{c|}{\textbf{MJOP}} & \multicolumn{4}{c|}{\textbf{AVAM}} & \multirow{2}{*}{\textbf{\textbf{NSR}}} & \multirow{2}{*}{\textbf{TMS}} & \multirow{2}{*}{\textbf{MAS}} & \multicolumn{3}{c|}{\textbf{MJOP}} & \multicolumn{4}{c|}{\textbf{AVAM}} & \multirow{2}{*}{\textbf{NSR}} & \multirow{2}{*}{\textbf{TMS}} & \multirow{2}{*}{\textbf{MAS}} & \multicolumn{4}{c|}{\textbf{AVAM}} \\ \cline{5-11} \cline{15-21} \cline{25-28}
                  &                      &                      &                      & $S_a$    & $S_b$   & $S_c$   & $R_a$ & $R_b$ & $R_c$ & $R_d$ &                      &                      &                      & $S_a$    & $S_b$   & $S_c$   & $R_a$ & $R_b$ & $R_c$ & $R_d$ &                      &                      &                      & $R_a$ & $R_b$ & $R_c$ & $R_d$ \\ \hline
\normalsize \textbf{CIFAR10}           & 60.2                 & 60.6                 & 48.7                 & 55.0    & 55.1   & 51.8   & 60.4 & 60.5 & 60.7 & 59.5 & 65.3                 & 64.9                 & 38.8                 & 55.7    & 51.8   & 26.3   & 64.1 & 64.6 & 65.0 & 64.2 & 83.7                 & 82.8                 & 69.6                 & 82.7 & 82.7 & 82.7 & 81.3 \\ \hline
\normalsize \textbf{MNIST}             & 98.8                 & 98.8                 & 98.2                 & 98.2    & 98.2   & 98.2   & 98.7 & 98.7 & 98.7 & 98.6 & 99.1                 & 98.7                 & 98.3                 & 97.9    & 98.0   & 97.2   & 98.8 & 98.8 & 98.8 & 98.9 & 99.3                 & 99.4                 & 98.9                 & 99.3 & 99.4 & 99.4 & 99.4 \\ \hline
\normalsize \textbf{FMNIST}            & 91.0                 & 90.5                 & 87.3                 & 88.0    & 88.1   & 88.0   & 90.2 & 90.2 & 90.3 & 89.4 & 89.9                 & 89.5                 & 79.5                 & 86.4    & 86.3   & 85.8   & 89.0 & 89.2 & 89.1 & 88.3 & 93.4                 & 93.2                 & 91.6                 & 93.2 & 93.3 & 93.4 & 93.2 \\ \hline
\end{tabular}
\caption{\textit{Accuracy Results (\%)}. \textbf{NSR}: \textbf{N}on-\textbf{S}piking \textbf{R}eLU ANN's results; \textbf{TMS}: \textbf{T}rue-\textbf{M}ax $U$ \textbf{S}NN's results; \textbf{MAS}: \textbf{M}ax-to-\textbf{A}vg \textbf{S}NN's results; \textbf{MJOP} \& \textbf{AVAM}: Spiking-MaxPooling SNN's results. For \textbf{CIFAR10} $S_a, S_b, S_c$ ($scale$) values are $1.0, 2.0, 5.0$; for \textbf{MNIST} they are $1.0, 1.2, 2.0$; and for \textbf{FMNIST} they are $1.0, 1.5, 2.0$ resp. $R_a, R_b, R_c, R_d$ ($radius$) values $\forall$ the datasets are $0.20, 0.25, 0.30, 1.0$ resp. As expected, $R_d = 1.0$ results in accuracy loss, since difference of $U_i$ is not always $\approx$ 1.0; thus, poor approximation of max $U_i$. Note that the SNNs do not necessarily outperform the ANNs because they are obtained after the ANN-to-SNN conversion.}
\label{tab:res}
\end{table*}

\section{Experiments \& Results}
We next describe the experiments conducted with the MJOP and AVAM methods of spiking-MaxPooling. In accordance with the ANN-to-SNN conversion paradigm, we first train a rate-neuron (\texttt{ReLU}) based model and then convert it to a spiking network (of IF spiking neurons). We use the NengoDL library \cite{rasmussen2019nengodl} for training, conversion, and inference; and the NengoLoihi library for deploying the SNNs on the Loihi boards. While training the models we ensure that they are properly tuned to account for the firing-rate quantization of IF spiking neurons. In the converted SNNs, we do the MaxPooling operation via our proposed methods of spiking-MaxPooling Nets. We use the MNIST, FMNIST, and CIFAR10 datasets (normalized $[-1, 1]$) and conduct experiments with $3$ different architectures (Fig. \ref{fig:archs}). For simplicity, we fix the MaxPooling window to $2 \times 2$ in all the architectures.
\subsection{Common settings in MJOP and AVAM methods}
For our proposed spiking-MaxPooling methods, each Conv-channel is flattened and the $2\times2$ pooling window inputs are grouped in sizes of $4$, and passed next to the layer of MJOP Nets or AVAM Nets depending on the choice of spiking-MaxPooling. The outputs from the individual Nets are collected in a channel wise manner and passed to the next Conv layer. 
While flattening the channels and passing the pooled inputs to the spiking-MaxPooling layer, and subsequently while collecting the outputs, one has to properly arrange the inputs and outputs to preserve the $2 \times 2$ MaxPooling layout. Note that the MJOP and AVAM methods are compatible with any ordering of the preceding Conv layer's channels.

\subsection{Model Details}
The first Conv layer in each architecture (in Fig. \ref{fig:archs}) has a kernel size of $(1, 1)$ which acts a pixel-value to spike converter. For training each architecture, we use the \texttt{Adam} optimizer with a learning rate of $1e-3$ and a decay of $1e-4$, and use the \texttt{Categorical Crossentropy} loss function with logits. For MNIST, Architectures $1$, $2$, and $3$ were trained for $8$ epochs each; for CIFAR10, $1$ and $2$ were trained for $64$ epochs each, and $3$ for $164$ epochs; and for FMNIST, $1$ and $2$ were trained for $24$ epochs each, and $3$ for $64$ epochs. The training was done on the NVIDIA Tesla P100 GPUs. During inference with SNNs, the individual test images (irrespective of the dataset) were presented for $50$, $60$, and $120$ time-steps to the Architectures $1$, $2$, and $3$ respectively. Code is \href{https://github.com/R-Gaurav/SpikingMaxPooling}{\textcolor{blue}{public}}.

\subsection{SNNs with MJOP Net based spiking-MaxPooling}
Since MJOP method is Loihi dependent (and not supported on the GPUs), we execute the converted SNNs right on the Loihi boards in inference mode. Fig. \ref{fig:max_jop_mp} shows an example of how MJOP spiking-MaxPooling can be done in any arbitrary architecture. The individual channels of the preceding Conv layer are flattened and mapped to a Neuro-Core each. Each Neuro-Core has an $Ensemble$ of MAX join-Op configured MC neurons (with $4$ compartments each as shown in Fig. \ref{fig:mc_neuron1}) to do the $2\times2$ MaxPooling. The outputs (quartered in length) from each $Ensemble$ deployed on the individual Neuro-Cores are collected and reshaped as individual channels, and then fed to the next Conv layer. This process is repeated for any additional MaxPooling layers in the network. While deploying the architectures on the Loihi boards, the first Conv layer and the last ``Output'' layer are executed Off-Chip; rest of the layers run On-Chip. Inference is done over the entirety of each dataset for varying $scale$ values. Due to the Loihi hardware resource constraints and no support for \texttt{same} padding (in Conv layers) in NengoLoihi (1.1.0.dev0), we were unable to execute Architecture $3$ on the Loihi boards. Table \ref{tab:res} shows the accuracy results for Architectures $1$ and $2$ - both run on Loihi.

\subsection{SNNs with AVAM Net based spiking-MaxPooling}
AVAM method of spiking-MaxPooling is hardware independent. Although the individual AVAM Nets with varying radii were executed on Loihi (Fig. \ref{fig:max_jop_poc}b), executing SNNs with them on the Loihi boards gets challenging due to the AVAM Net layers exceeding the maximum supported number of input and output axons on the Loihi hardware. Therefore, we execute the SNNs with AVAM method of spiking-MaxPooling on GPUs only. Similar to the case of MJOP Net based SNNs, the individual channels in the preceding Conv layer are flattened and the pooling window's grouped values (in sizes of $4$) are passed to the layer of AVAM Nets to estimate the max values; which are then collected, reshaped as channels, and forwarded next. This process is repeated for any additional MaxPooling layers in the network. In the experiments with each of the three architectures, $\phi$ is set to $250$Hz, and the $radius$ value is varied; Table \ref{tab:res} shows the accuracy results.

\section{Discussion}
\subsection{Table \ref{tab:res} Result Analysis}
To evaluate the efficacy of SNNs with our proposed methods of spiking-MaxPooling, we compare it against the baseline results obtained with the ``True Max $U$'' based SNNs (column TMS) and its \texttt{ReLU} based non-spiking counterpart (column NSR, with TensorFlow) - both with regular $max()$ based MaxPooling, as well as with the SNNs where MaxPooling layers are replaced with AveragePooling layers after training (column MAS, only $Ensemble$s and associated connections removed in AVAM Net). In case of MNIST, the MJOP based SNNs perform similar to their non-spiking counterpart and the ``True Max $U$'' based SNN across all the architectures. In case of FMNIST, the performance drop of MJOP based SNNs is noticeable; and in case of CIFAR10, they perform very poorly. One reason for the accuracy drop is the Loihi hardware constraints i.e. $8$-bit quantization of the network weights and fixed-point arithmetic. We found the other reason in case of CIFAR10 to be the highly varying ISI distribution (of Conv layer neurons) across the test images -- possibly due to color channels (resulting in highly variable neuron activations). This prevented the choice of a $scale$ value to generalize well across all the test images. For MNIST (and to a large extent for FMNIST), we found the ISI distribution of Conv layer neurons to be light-tailed and mostly similar across the test images. More details on the ISI distribution in the Technical Appendix/Supplementary Material section below. In a separate experiment, setting two different $scale$ values $(2 \text{ and } 1.5)$ for the corresponding MaxPooling layers in Architecture $2$ for FMNIST did not improve the results. Overall, although the MJOP based SNNs were successfully deployed on Loihi, the MJOP spiking-MaxPooling seems suboptimal due to its poor generalizability. For its optimal performance, the maximally firing neuron in each pooling window (in a preceding Conv layer) should have (nearly) same ISI for a consistent effect of the $scale$ value. On the other hand, AVAM based SNNs perform at par with their non-spiking counterpart and the ``True Max $U$'' based SNN across all the three datasets and architectures. An important distinction between the MJOP \& AVAM methods is that the MJOP method estimates the ``True Max $U$'' indirectly by scaling the $U_{out}$ (note that this $U_{out}$ \textit{corresponds} to the maximum input $U_i$), whereas the AVAM method estimates the ``True Max $U$'' directly from the group of $U_i$, which makes the AVAM spiking-MaxPooling more robust and effective; and its optimal performance with the same set of $radius$ values across all the three datasets and architectures promises its generalizability. It is interesting to note that in some cases, AVAM based SNNs perform better than their non-spiking counterpart and/or ``True Max $U$'' based SNNs. Also, the SNNs with AveragePooling layers (column MAS) perform poorly compared to all other networks.

\subsection{Adapting MJOP \& AVAM spiking-MaxPooling}
Our proposed methods of spiking-MaxPooling are suitable for the rate-based SNNs which represent activations as currents (i.e. filtered/synapsed spikes). SNNs which do not filter the spike trains and work directly on binary spikes \cite[\dots]{sengupta2019going, han2020deep}, cannot adapt our proposed methods; this holds true for Time-To-First-Spike based SNNs as well. With respect to the scalability of the MJOP and AVAM Net against the size of pooled inputs, it is linear. For an $r\times c$ MaxPooling window (where $r, c \in \mathbb{N}$), the number of compartments/neurons required in the MJOP and AVAM Net is $r \times c$ and $2\times (r\times c) - 2$ respectively. However, with the increase in the number of neurons in the hierarchical AVAM Net, the estimated max output may get noisier; although, this doesn't hold true for MJOP based method. AVAM Net based spiking-MaxPooling is deployable on any neuromorphic hardware that supports weighted connections and spiking neurons. In our MJOP based SNNs experiments, each channel of a Conv layer (prior to a MaxPooling layer) was small enough in dimensions such that the flattened vector was of size $\leq 1024$, thus easily mapped to a Neuro-Core. If the channel's dimensions are sufficiently large and the flattened vector's size is $> 1024$, then one needs to spatially split the channel and properly map it to more than one Neuro-Core such that no pooling window (thus the corresponding MC neuron) spans across two or more Neuro-Cores, as Loihi restricts the creation of a MC neuron on one Neuro-Core only. However, such a procedure need not be followed with the AVAM based SNNs (if deployed on GPUs).

\section{Conclusion}
To our best knowledge, this work is a first to present two different hardware-friendly methods of spiking-MaxPooling operation in SNNs, with their evaluation on Loihi. In a first, we also deployed SNNs with MaxPooling layers (via MJOP method) on the Loihi boards. Note that our goal was not to outperform the SoTA results on the experimented datasets, rather to show the efficacy of our hardware-friendly spiking-MaxPooling methods, which we do so by achieving comparable results with the regular MaxPooling based ANNs (column NSR in Table \ref{tab:res}). For the appropriate choice of the tunable $scale$ and $radius$ values in MJOP \& AVAM Net respectively, we also presented a heuristic method. The Proof-of-Concept Demonstrations of both the methods on Loihi show that our work makes successful strides towards providing a solution to the MaxPooling problem in SNNs. This also opens up avenues for building hardware ready SNNs with MaxPooling layers without compromising on the quality (otherwise due to replacing MaxPooling with AveragePooling). One immediate future direction of our work is to enable the deployment of AVAM based SNNs on a neuromorphic hardware. Next, we hope that our work encourages efforts towards developing other hardware-friendly methods of spiking-MaxPooling.
\section*{Acknowledgment}
We would like to thank Eric Hunsberger and Xuan Choo (from \href{https://appliedbrainresearch.com}{\textcolor{blue}{ABR}}) for their help with the NengoLoihi and NxCore details to deploy our spiking-MaxPooling methods on Loihi.

\bibliographystyle{./IEEEtran}
\bibliography{./IEEEabrv,./IEEEexample}

\newpage

\onecolumn

\section*{\textbf{TECHNICAL APPENDIX / SUPPLEMENTARY MATERIAL}}

\vspace{10pt}

\begin{appendices}

This appendix contains two sections: \textbf{Appendix \ref{sec:1}} corresponds to the ISI distributions plots and \textbf{Appendix \ref{sec:2}} corresponds to the details of experiment parameters for reproducibility.

\vspace{10pt}
\section{ISI DISTRIBUTION PLOTS \label{sec:1}}

\begin{figure}[]
    \centering
    \includegraphics[scale=0.5325]{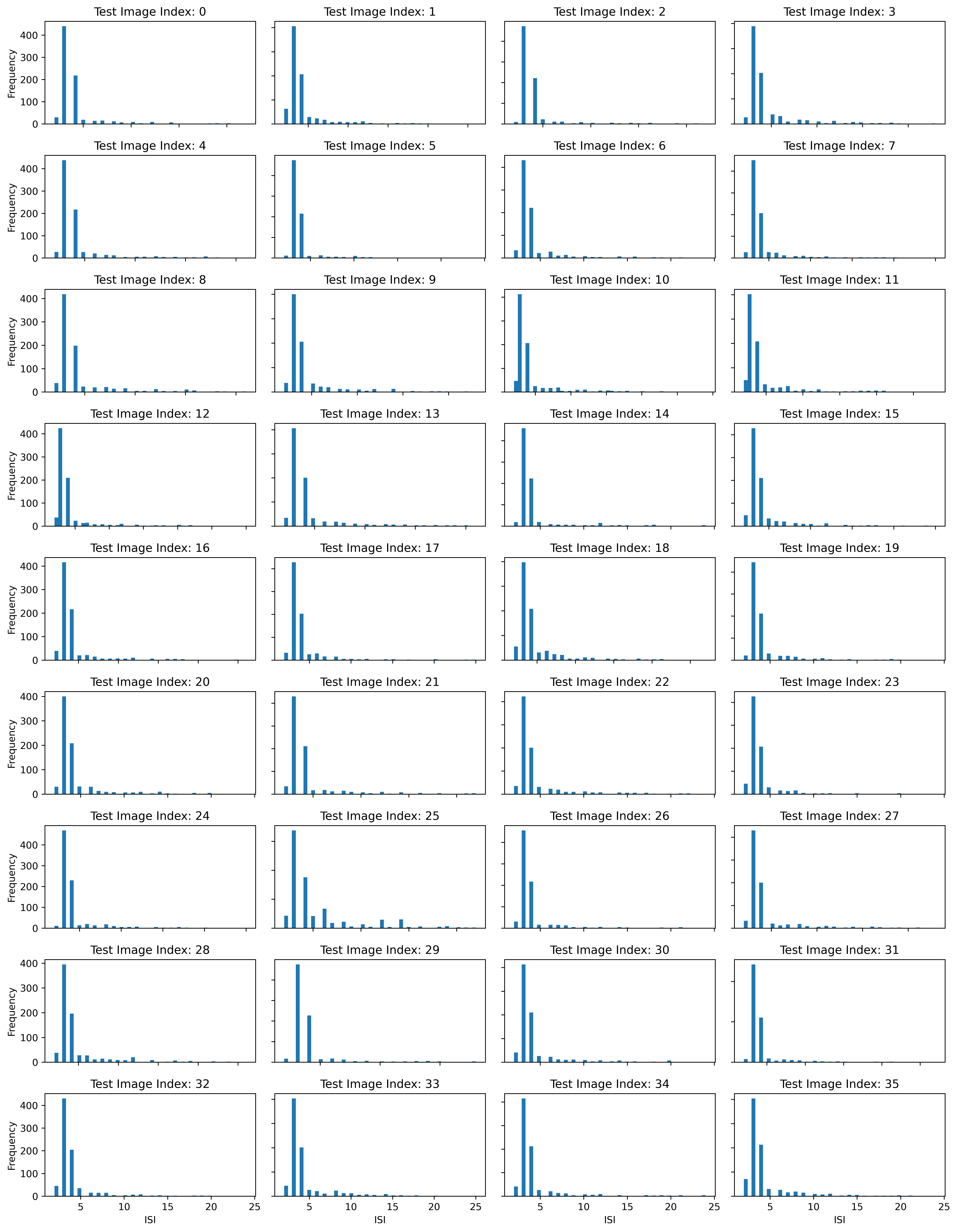}
    \vspace{-10pt}
    \caption{\textit{Architecture $1$, MNIST, Conv\_1}. ISI distributions are light-tailed and similar.}
    \label{fig:m1_mn_conv1}
\end{figure}

\begin{figure}[]
    \centering
    \includegraphics[scale=0.5325]{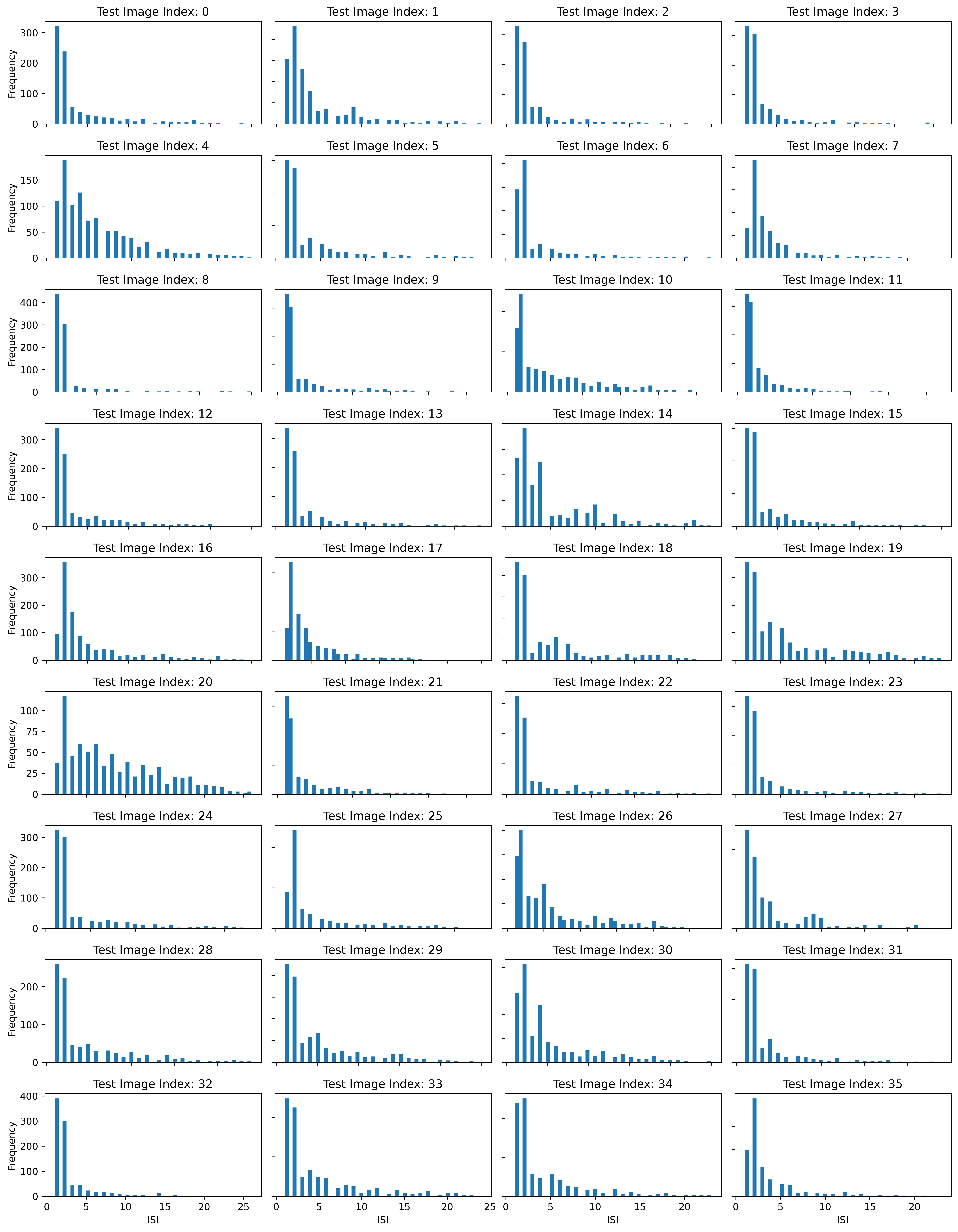}
    \vspace{-10pt}
    \caption{\textit{Architecture $1$, FMNIST, Conv\_1}. Most ISI distributions are light-tailed and mostly similar.}
    \label{fig:m1_fmn_conv1}
\end{figure}

\begin{figure}[!h]
    \centering
    \includegraphics[scale=0.5325]{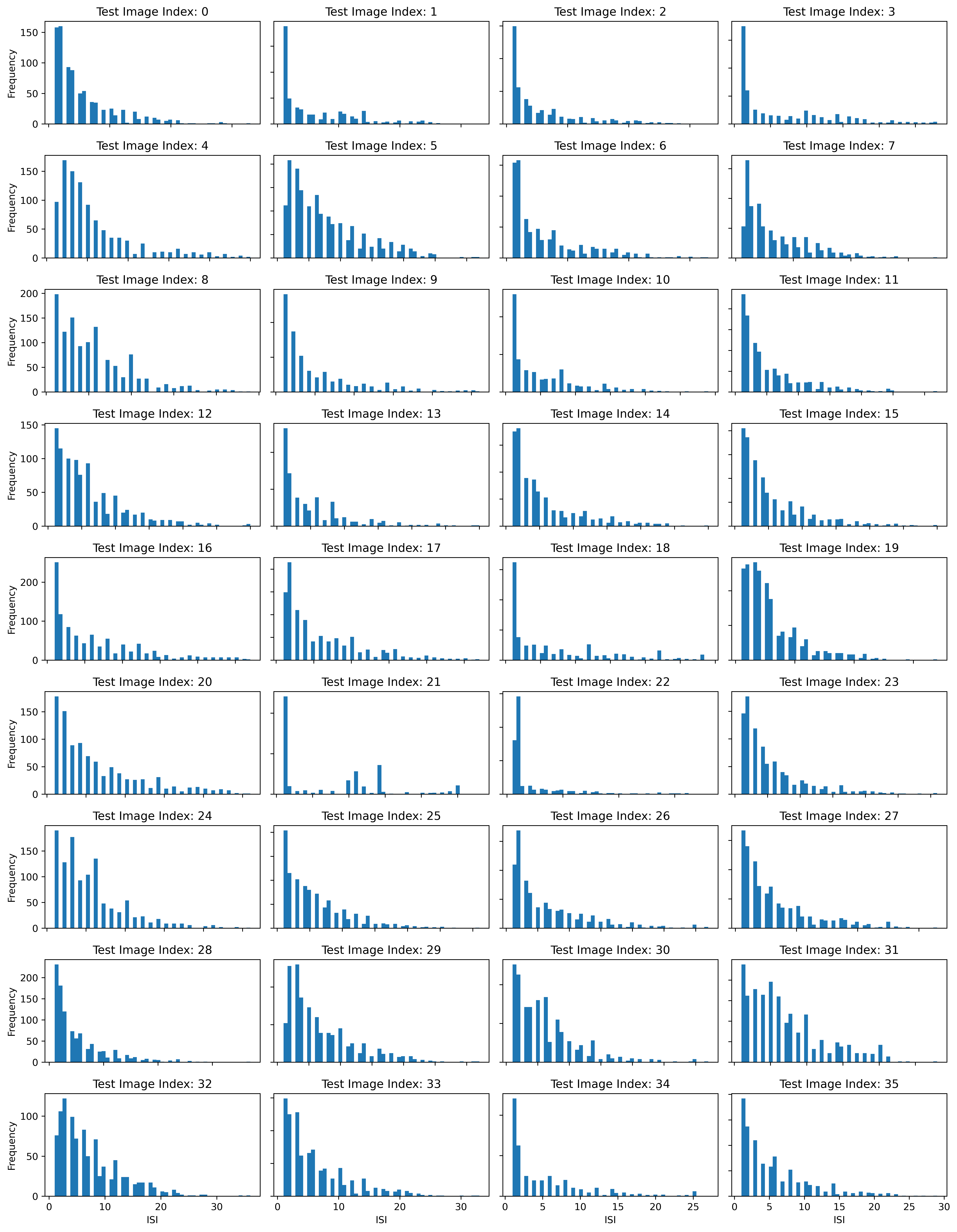}
    \vspace{-10pt}
    \caption{\textit{Architecture $1$, CIFAR10, Conv\_1}. ISI distributions are dissimilar and mostly fat-tailed.}
    \label{fig:m1_c10_conv1}
\end{figure}

\begin{figure}[!h]
    \centering
    \includegraphics[scale=0.5325]{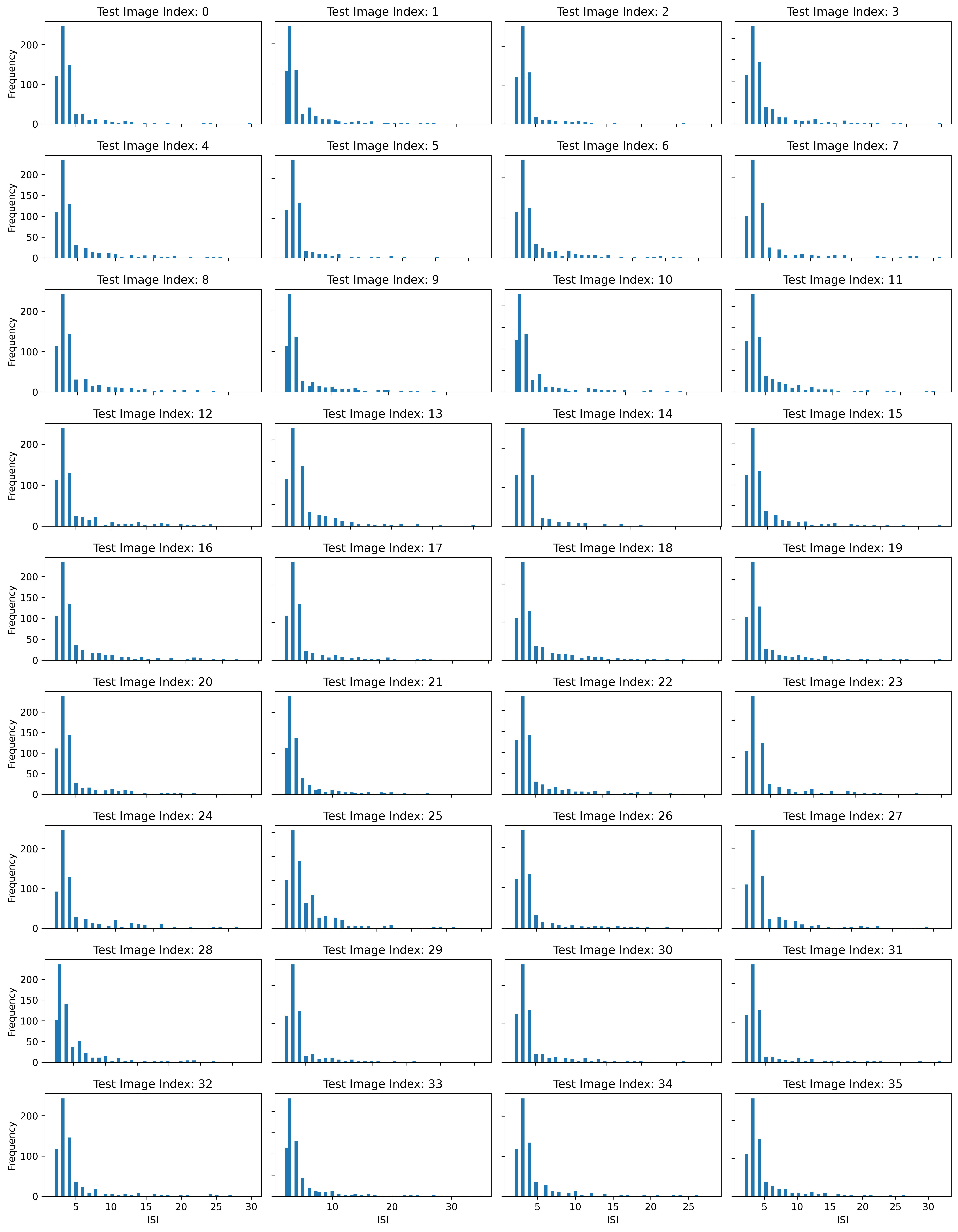}
    \vspace{-10pt}
    \caption{\textit{Architecture $2$, MNIST, Conv\_1}. ISI distributions are light-tailed and similar.}
    \label{fig:m2_mn_conv1}
\end{figure}

\begin{figure}[!h]
    \centering
    \includegraphics[scale=0.5325]{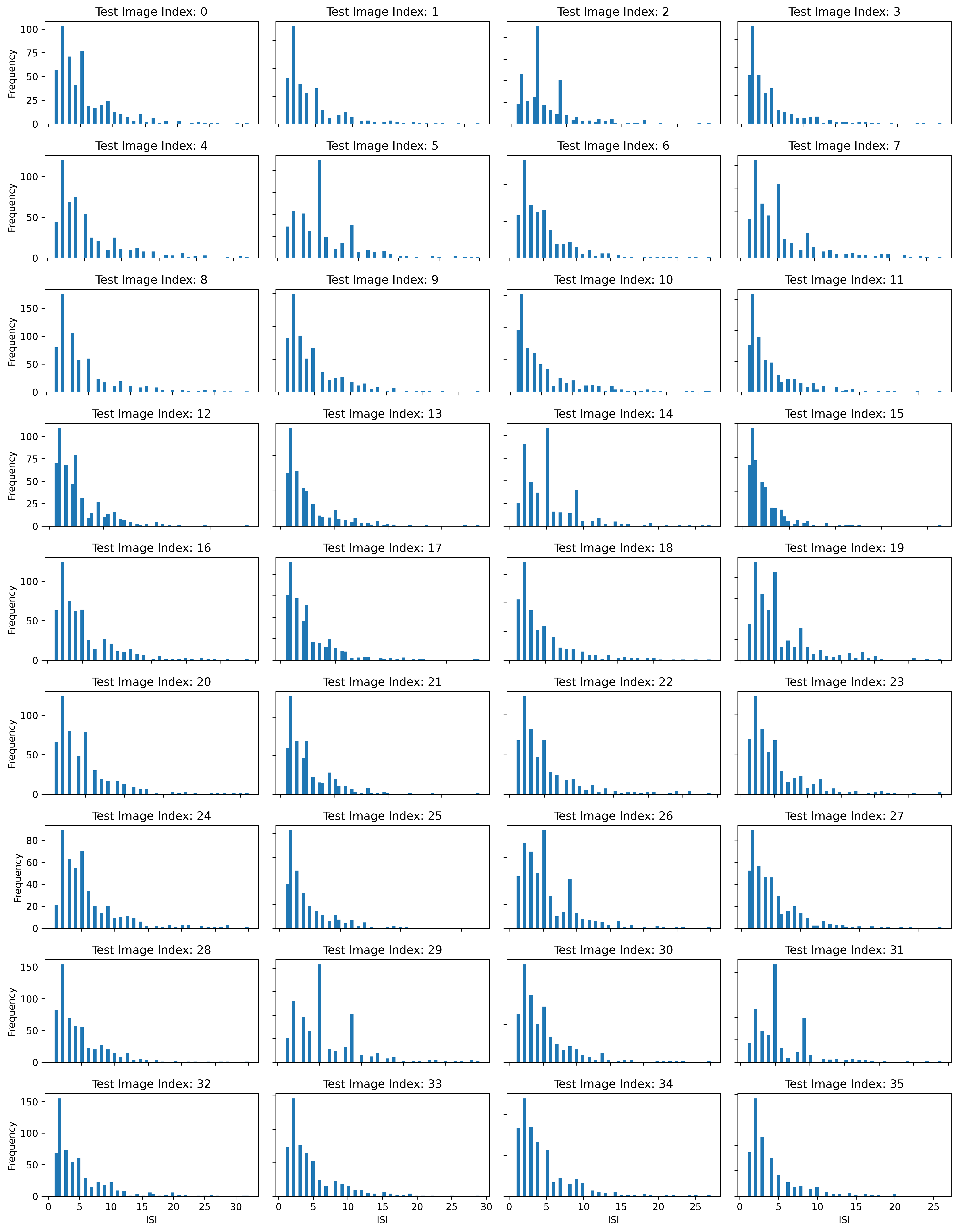}
    \vspace{-10pt}
    \caption{\textit{Architecture $2$, MNIST, Conv\_2}. ISI distributions are not very similar, and not very light-tailed.}
    \label{fig:m2_mn_conv2}
\end{figure}

\begin{figure}[!h]
    \centering
    \includegraphics[scale=0.5325]{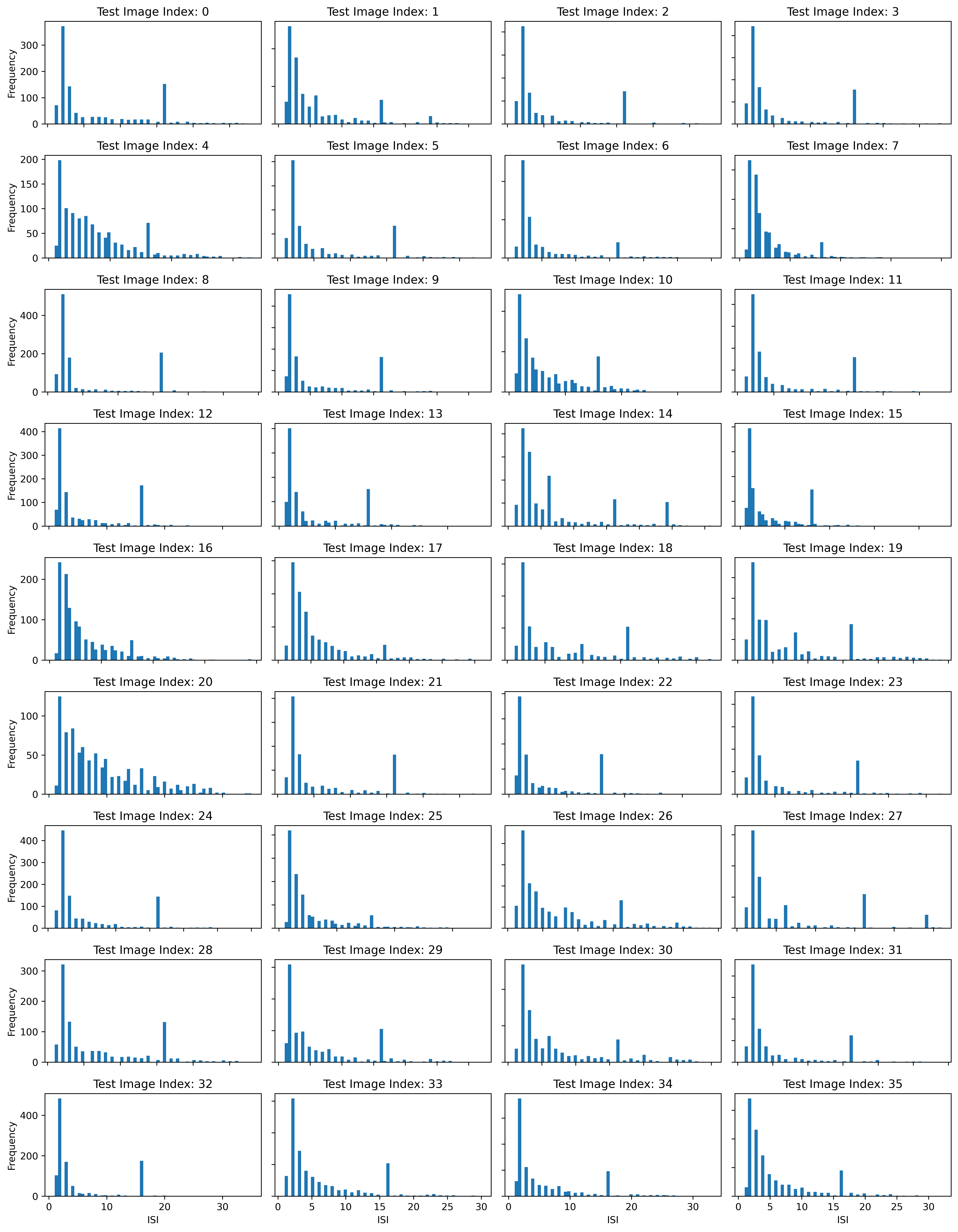}
    \vspace{-10pt}
    \caption{\textit{Architecture $2$, FMNIST, Conv\_1}. ISI distributions are almost light-tailed and mostly similar.}
    \label{fig:m2_fmn_conv1}
\end{figure}

\begin{figure}[!h]
    \centering
    \includegraphics[scale=0.5325]{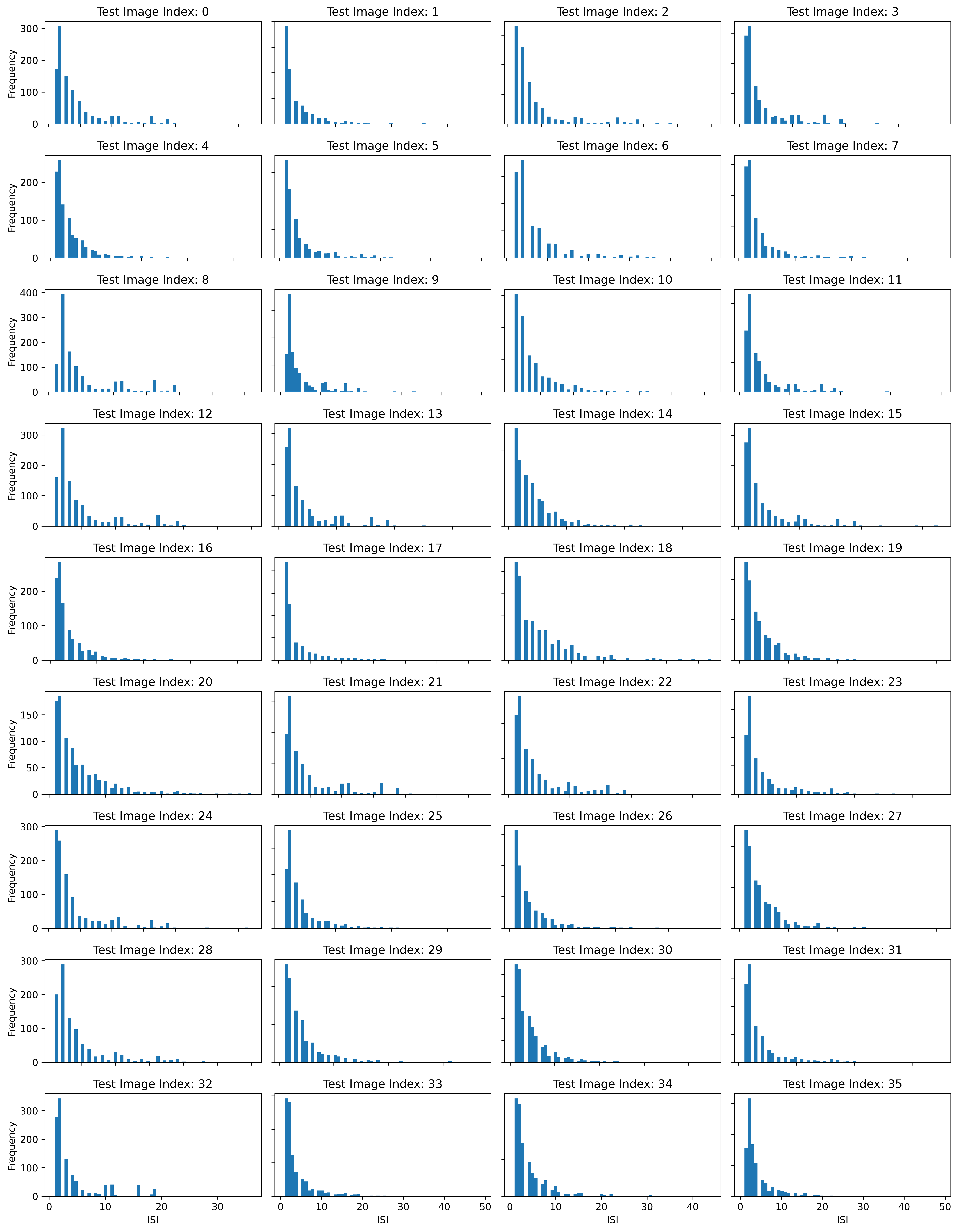}
    \vspace{-10pt}
    \caption{\textit{Architecture $2$, FMNIST, Conv\_2}. ISI distributions are light-tailed and similar.}
    \label{fig:m2_fmn_conv2}
\end{figure}

\begin{figure}[!h]
    \centering
    \includegraphics[scale=0.5325]{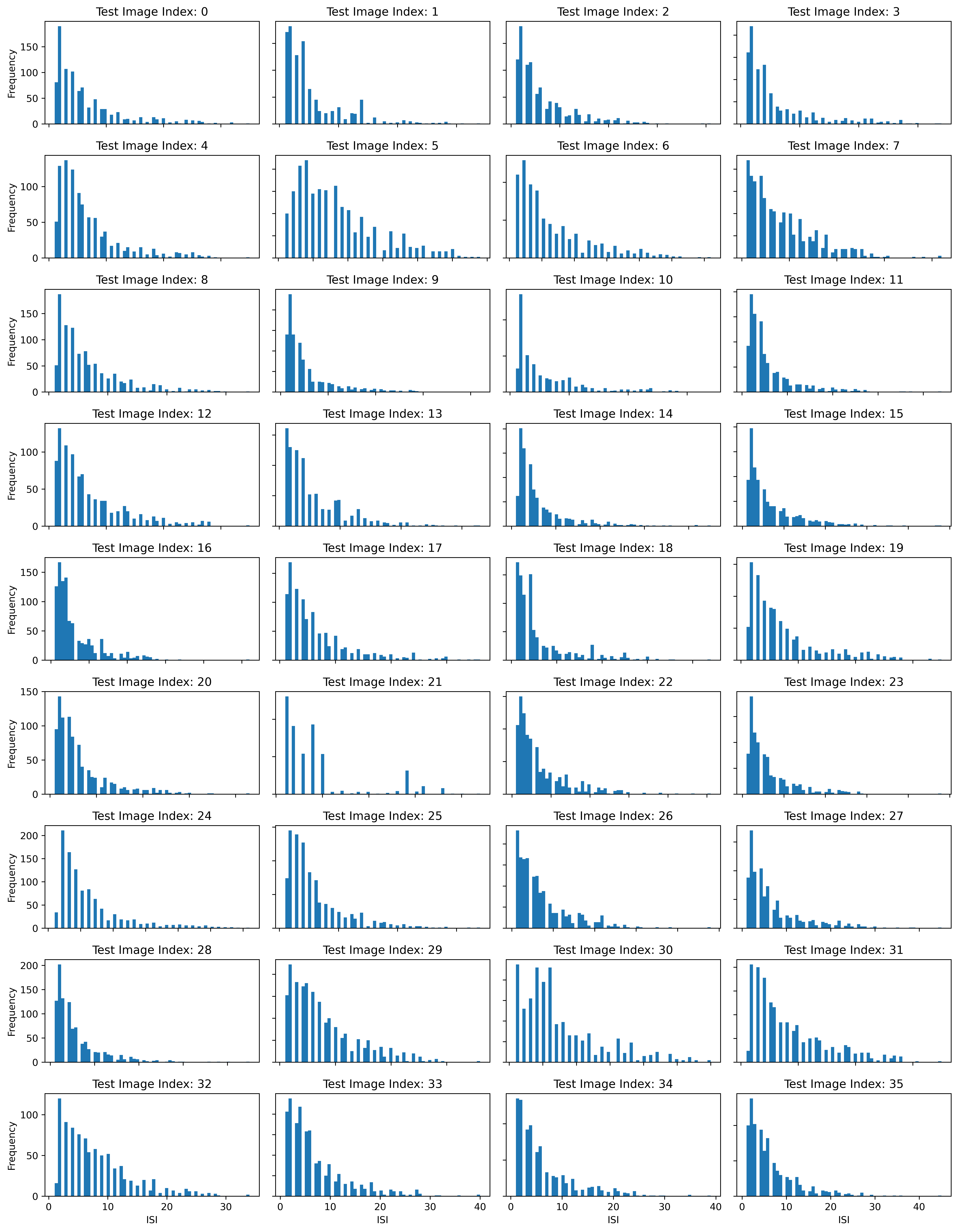}
    \vspace{-10pt}
    \caption{\textit{Architecture $2$, CIFAR10, Conv\_1}. ISI distributions dissimilar and fat-tailed.}
    \label{fig:m2_c10_conv1}
\end{figure}

\begin{figure}[!h]
    \centering
    \includegraphics[scale=0.5325]{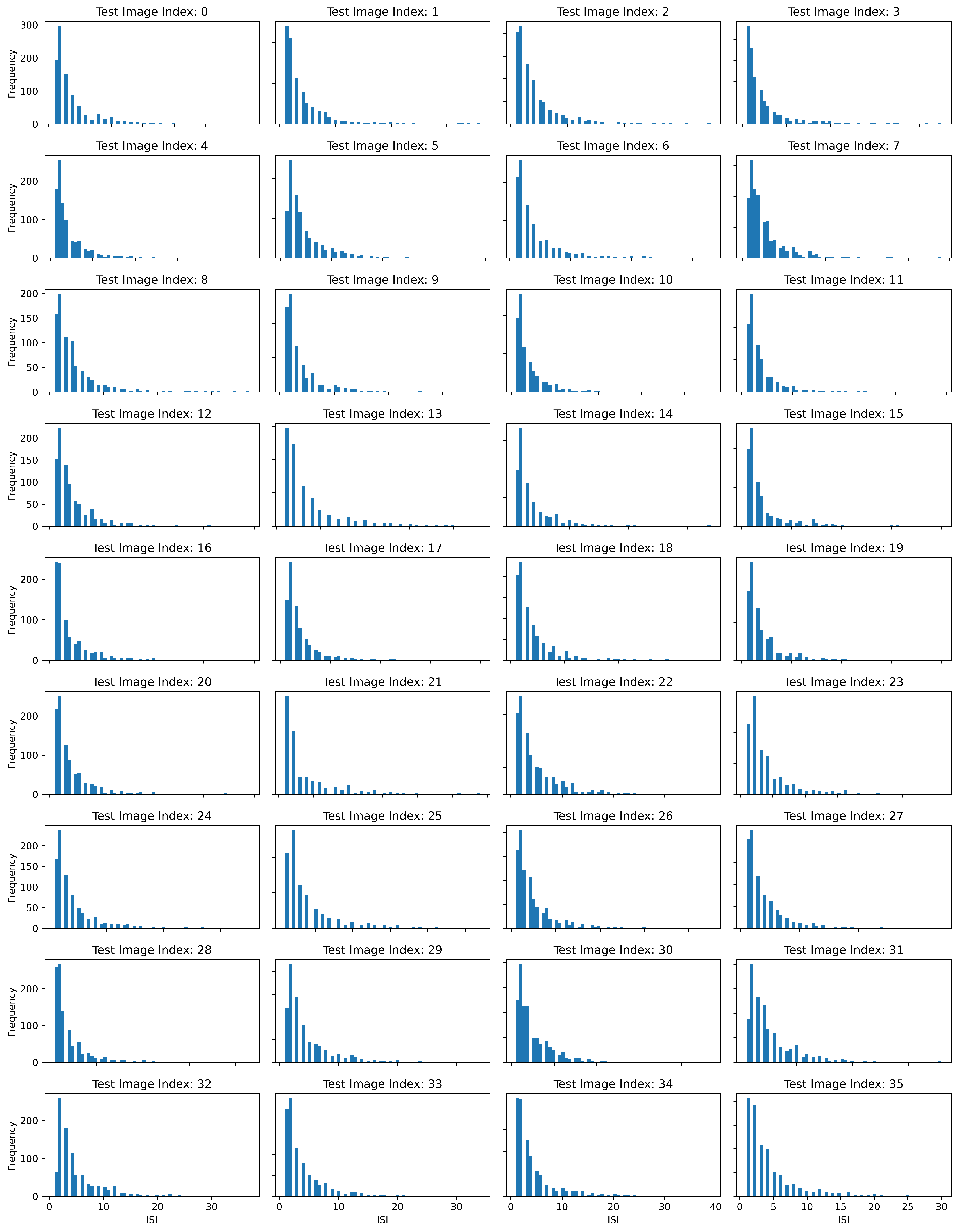}
    \vspace{-10pt}
    \caption{\textit{Architecture $2$, CIFAR10, Conv\_2}. ISI distributions are light-tailed and similar.}
    \label{fig:m2_c10_conv2}
\end{figure}

Herein, we plot the ISI distributions of the Conv layer neurons (random and $2048$ in number) of $36$ test images of all the three datasets each with respect to the Architectures $1$ and $2$, to support our claims in the paper. The plots correspond to the Conv layers prior to the MaxPooling layers only and are obtained from ``True Max $U$'' based SNNs. As mentioned in the paper, for the optimal performance of MJOP Net, the maximally firing neuron in each pooling window should have nearly same ISI for a consistent effect of the $scale$ value. If the ISIs of maximally firing neurons are very different, a single $scale$ value doesn't generalize well (as the $scale$ value is dependent on $U_i$).
\\\\
Therefore, for the datasets where ISI distributions are light-tailed and mostly similar across the test images, a suitable choice of $scale$ value generalizes well (across the test images), thus desirable performance of MJOP based SNNs. Light-tailed distribution implies that most of the neurons fire at lower ISIs and those ISIs are nearly same (as will be seen in the case of MNIST and to a large extent in FMNIST dataset), thus a high probability of (nearly) same ISI of maximally firing neurons in MaxPooling windows. Datasets for which the distributions are fat-tailed and are varied across the test images (as will be seen in case of CIFAR10 dataset), a chosen $scale$ value doesn't generalize well for them; thus making it difficult to tune the $scale$ parameter. Moreover, in case of a deeper architecture, a poor approximation of an earlier MaxPooling layer due to suboptimal $scale$ value further degrades the approximation of later MaxPooling layers.
\\\\
Fig. \ref{fig:m1_mn_conv1} to Fig. \ref{fig:m2_c10_conv2} are the ISI distribution plots for the three experimented datasets and two architectures. Note that Architecture $1$ has only one MaxPooling layer, and Architecture $2$ has two MaxPooling layers.

\vspace{10pt}

\section{Experiment Parameters in Detail \label{sec:2}}
\begin{itemize}
    \item To demonstrate the indifference of our spiking-MaxPooling methods to different channels ordering, we executed the experiments with following settings:
    \begin{itemize}
        \item For experiments with MJOP based SNNs, in each architecture, the ordering of Conv channels in case of MNIST, CIFAR10, and FMNIST was \texttt{channels\_last}, \texttt{channels\_first}, and \texttt{channels\_last} respectively.

        \item For experiments with AVAM based SNNs, in Architectures $1$ and $2$ each, the ordering of Conv channels for MNIST, CIFAR10, and FMNIST was \texttt{channels\_first}, \texttt{channels\_last}, and \texttt{channels\_last} respectively; in Architecture $3$, it was \texttt{channels\_last} for all the datasets.
    \end{itemize}

    \item While training the models with NengoDL Library, we set the neuron type in \texttt{nengo\_dl.Converter()} as \texttt{nengo\_loihi.neurons.LoihiSpikingRectifiedLinear()}. It still uses \texttt{ReLU} internally but with the added information to account for the quantization of firing rates and the set limit of $1$ spike per time-step on Loihi. We set the \texttt{scale\_firing\_rates} parameter to $400$. Training images are presented for one time-step only.

    \item While executing the converted SNNs in inference mode, we set the neuron type and \texttt{scale\_firing\_rates} same as above with \texttt{synapse} set to $0.005$.

    \item The configuration settings for MJOP Net (i.e. the $4$ compartmental neuron) are: \texttt{gain} set to $1000$, \texttt{bias} set to $0$, \texttt{neuron\_type} set to \texttt{nengo\_loihi.neurons.LoihiSpikingRectifiedLinear()} in which \texttt{amplitude} $=\frac{scale}{1000}$ and initial \texttt{voltage} $=0$, in \texttt{nengo.Ensemble()}.

    \item The configuration settings for AVAM Net (apart from the ones mentioned in the paper) are: \texttt{neuron\_type} set to \texttt{nengo\_loihi.neurons.LoihiSpikingRectifiedLinear()}, \texttt{encoders} set to $[1, -1]$, \texttt{intercepts} set to $[0, 0]$ in each $nengo.Ensemble$ (of two neurons). The connections from the $Ensemble$ neurons to the nodes have \texttt{synapse} set to $0.001$ and \texttt{transform} set to $\frac{radius}{\phi}$ where $\phi$ is the max firing rate mentioned in the paper.

    \item For more information on all the above parameters, documentations can be found at \texttt{\href{https://www.nengo.ai}{https://www.nengo.ai}}.

\end{itemize}

\newpage

\newpage

\newpage

\newpage

\newpage

\newpage

\newpage

\newpage

\newpage


\end{appendices}

\end{document}